\pdfoutput=1

\documentclass[11pt]{article}

\usepackage[final]{acl}

\usepackage{times}
\usepackage{latexsym}

\usepackage[T1]{fontenc}

\usepackage[utf8]{inputenc}

\usepackage{microtype}

\usepackage{inconsolata}

\usepackage{graphicx}
\usepackage{float}

\usepackage{enumitem}
\usepackage{amsmath}
\usepackage{subfig}
\usepackage{array}
\usepackage[table, dvipsnames]{xcolor}
\usepackage{multirow}
\usepackage{makecell}
\usepackage{tikz}
\usepackage{subfig}
\usepackage{booktabs}
\usepackage{tabularx}
\usepackage{subcaption}
\usepackage{fvextra}
\newenvironment{minted}[2][]{\VerbatimEnvironment\begin{Verbatim}[breaklines,breakanywhere,breaksymbolleft={},breaksymbolright={},fontsize=\scriptsize]}{\end{Verbatim}}

\usepackage{array}
\newcolumntype{Y}{>{\raggedright\arraybackslash}X}

\newcommand*\circled[1]{\tikz[baseline=(char.base)]{\node[shape=circle,draw,inner sep=1pt] (char) {\small #1};}}
\newcommand*\circledLett[1]{\tikz[baseline=(char.base)]{\node[shape=circle,draw,inner sep=1.5pt] (char) {\small #1};}}

\newcommand{\trip}{\textsc{(Subject, Relationship, Object)} }
\newlist{questions}{enumerate}{2}
\setlist[questions]{leftmargin=2.45em}
\setlist[questions,1]{label=\textbf{RQ\arabic*},ref=RQ\arabic*}
\setlist[questions,2]{label=(\alph*),ref=\thequestionsi(\alph*)}

\newif\ifrevisions
\revisionsfalse

\newcommand{\rev}[1]{\ifrevisions\textcolor{Green}{#1}\else#1\fi}

\newcommand{\revblock}[1]{%
  \ifrevisions{\begingroup\color{Green}#1\endgroup}%
  \else #1\fi
}

\title{Comparing Human and Large Language Model Interpretation of Implicit~Information}

\author{
Antonio De Santis \quad Tommaso Bonetti \quad Andrea Tocchetti \quad Marco Brambilla \\
Politecnico di Milano, Italy \\
\small{\textbf{Correspondence:} \href{mailto:antonio.desantis@polimi.it}{antonio.desantis@polimi.it}}
}

\begin{document}
\maketitle
\begin{abstract}
The interpretation of implicit meanings is an integral aspect of human communication. However, this framework may not transfer to interactions with Large Language Models (LLMs). To investigate this, we introduce the task of Implicit Information Extraction (IIE) and propose an LLM-based IIE pipeline that builds a structured knowledge graph from a context sentence by extracting relational triplets, validating implicit inferences, and analyzing temporal relations. We evaluate two LLMs against crowdsourced human judgments on two datasets. We find that humans agree with most model triplets yet consistently propose many additions, indicating limited coverage in current LLM-based IIE. Moreover, in our experiments, models appear to be more conservative about implicit inferences than humans in socially rich contexts, whereas humans become more conservative in shorter, fact-oriented contexts. Our code is available at \href{https://github.com/Antonio-Dee/IIE_from_LLM}{this URL}.
\end{abstract}

\section{Introduction}\label{sec:introduction}

\begin{figure}[t]
    \centering
    \includegraphics[width=0.7\linewidth]{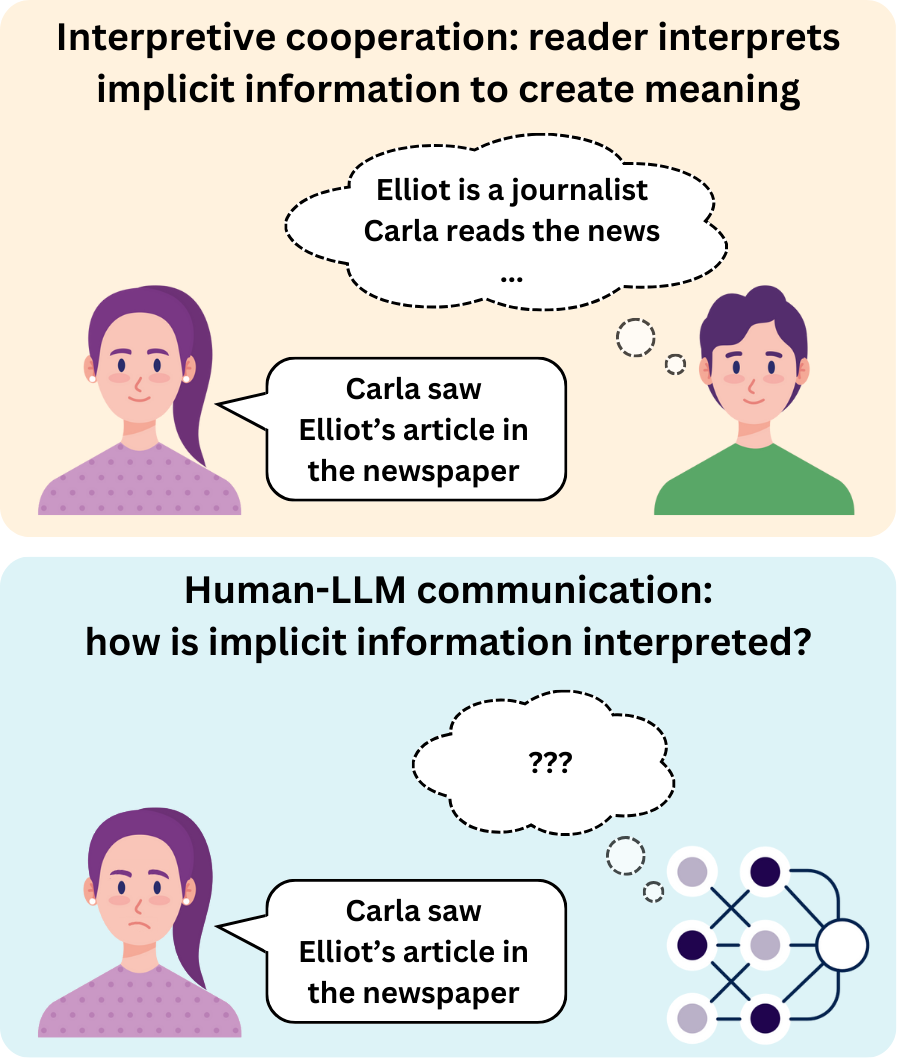}
    \caption{Comparison of the interpretation of implicit meaning in human and human-LLM communication.}
    \label{fig:human_comm}
\end{figure}

Large Language Models (LLMs) have revolutionized Natural Language Processing (NLP), performing remarkably well on a multitude of open problems \cite{LLMsIRsurvey, LLMsQAsurvey, LLMsClassSurvey} and becoming widely popular due to their ability to quickly generate text nearly indistinguishable from human-generated language \cite{LLMapps2}. LLM-generated content has therefore proliferated on the Web and in several professional fields \cite{AIwiki, LLMwriting, LLMsAcademia}. Humans, however, communicate within a specific framework that may not apply to language models \cite{parrots}. As semiotician Umberto Eco suggests, the meaning of human-generated text is created collaboratively, with the author and the reader engaging in the process of \textit{interpretive cooperation} \cite{Eco}. This theory assigns an active role to the reader in interpreting the implicit meaning intrinsic to any piece of text (see Fig. \ref{fig:human_comm}). Although humans tend to communicate within this framework, this interpretive and collaborative approach may not apply when a human is interacting with LLM-generated text. Therefore, we set about analyzing and comparing humans and LLMs in terms of how they interpret implicit information.

Current research regarding information extraction with LLMs is rich, but lacks focus on implicit information. To bridge this gap, we propose an Implicit Information Extraction (IIE) approach that automatically builds a structured representation of the model’s understanding of a context sentence. Our method can be applied to black-box language models, as it does not require access to the model parameters or fine-tuning. The result is a knowledge graph (KG) with two tiers, one for relational information and the other for temporal data. The LLMs’ knowledge graphs are then compared with human answers regarding the same context sentences, involving both direct comparison and agreement questions to provide an informative comparison that considers both quantitative and qualitative aspects. The analysis is driven by four research questions: \textbf{(RQ1)} What are the main \textit{drivers} of inference for LLMs and humans? \textbf{(RQ2)} How do LLMs compare to humans in terms of \textit{strictness} when evaluating potential inferences? \textbf{(RQ3)} Can LLMs effectively parse \textit{timing} relations between events? \textbf{(RQ4)} How strong is the \textit{consensus} among humans approaching this task?

\section{Background and Related Work}\label{sec:background}

\paragraph{Implicit Information Extraction}

Implicit information (implicit meaning) is defined as the distinction between \textit{“what a text says (\textit{i.e.}, its explicit or literal meaning)”} and \textit{“what is inferable from the text (\textit{i.e.}, its implicit meaning)”} \cite{ImplicitMeaningDef}. To our knowledge, there exists little research specific to the extraction of implicit information from text. However, LLMs have recently demonstrated the ability to effectively derive some implicit information through in-context learning for question answering \cite{implicitREforQA} and relation extraction using fine-tuning and retrieval-augmented generation \cite{RAG+RE}.  
\rev{From a different perspective, inferable content is often modeled through Natural Language Inference (NLI), where a model judges whether a premise sentence entails a hypothesis sentence. This entailment-based notion is stricter than the definition of implicit meaning we adopt from \citet{ImplicitMeaningDef}, since this notion includes inferable content beyond strict entailment and our goal is to extract an open set of structured triplets rather than to output a discrete label.}

\paragraph{Open Information Extraction}

Open Information Extraction (OIE) is a broad task whose goal is extracting information from text in the form of \trip triplets. Standard information extraction is typically applied using a pre-defined schema of relation types, making it suitable for analyzing domain-specific text (\textit{e.g.}, medical records). On the other hand, OIE removes the schema constraint, allowing it to effectively deal with unstructured text that features unseen relationships. LLMs have proven to be well-suited for this task, eliminating the need for supervised training due to the generalization capabilities developed during pre-training \cite{OIEsurvey}. Even in a zero-shot setting, these models have been able to match the performance of state-of-the-art supervised methods, although hallucinations are a known problem affecting their accuracy \cite{HallucinationSurvey, LLMapps1}. Techniques such as few-shot prompting and fine-tuning, however, have proven capable of mitigating this and other issues to some extent \cite{OIEwLLMs}.

\paragraph{Knowledge Graph Extraction}

Since OIE extracts relational \trip triplets, it is possible to build a structured representation of a set of such triplets using a knowledge graph (KG). A KG is a formalism that uses nodes and edges to represent entities and their respective relationships. The entities are the subjects and objects of the extracted triplets, while each relationship edge represents one triplet \cite{KGstory}. Knowledge graph extraction is the process of extracting and organizing triplets in a KG. This procedure can be applied, for instance, to generate a structured knowledge model from text \cite{KGCsurvey}. Furthermore, KGs have the advantage of facilitating comparison and integration due to their well-defined structure.

Knowledge graph extraction using LLMs is often employed as a means of eliciting the intrinsic (or parametric) knowledge of a model. Some studies focus on entities and relationships \cite{internalKB}, while others investigate the ontological or commonsense knowledge of LLMs \cite{Crum}. For instance, \citet{KnowledgeDistil} instruct a target model to generate inferences starting from a given event based on the taxonomy of relation types provided by \textsc{Atomic} \cite{Atomic}. These relation types can be useful in the extraction and analysis of implicit information in general, as they cover the inference of other events, mental states, and personal attributes. Lastly, part of the existing research seeks to build knowledge graphs from text in a semantically-aware fashion \cite{EDC,tasi}. Indeed, taking semantics into consideration is crucial when building a KG from general-purpose text, where no pre-defined schema exists.


A similar and related task is the extraction of temporal knowledge graphs. In this case, the nodes of the graph represent events and the edges represent temporal relationships. This type of KG conveys the absolute and relative temporal position of events, each of which can be represented as a relational triplet in its own right \cite{EventKGsurvey}. After the events are extracted, they can undergo different analysis steps to reveal their position in time \cite{RospocherEvents} and duration \cite{EventPlus} as well as relationships with other events \cite{AllenIntervals, EventPlus, TimeML}. Additional aspects include the polarity (\textit{i.e.}, positive/negative) and the modality (\textit{i.e.}, asserted/hypothetical) of extracted events \cite{EventPlus, ModalityTimeML}. A relevant work in this field is TimeML \cite{TimeML}, a specification language dedicated to describing events and temporal data that covers many of the aspects described above.

\section{Methodology}

Given the lack of tailored implicit information extraction methods in the literature, we propose a methodology specifically designed to extract implicit meaning and compare the results with human knowledge models. Given the prominent role of personal interpretation, we want to capture as much information as possible about how LLMs perceive language by focusing on open-ended extraction, allowing a deeper qualitative analysis of similarities and differences between the knowledge models of humans and LLMs.

\begin{figure}[ht]
    \centering
    \includegraphics[width=.9\linewidth]{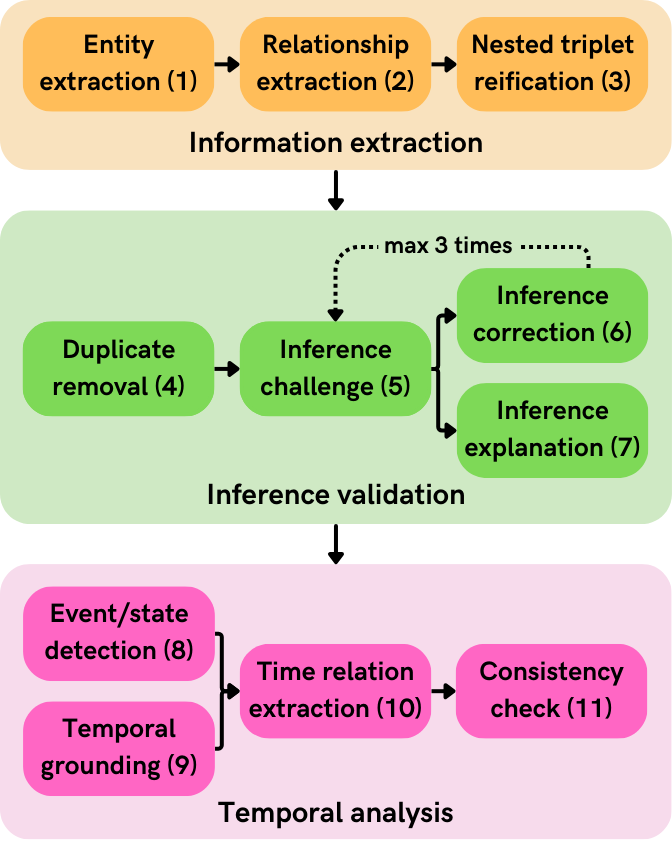}
    \caption{The proposed solution pipeline and its stages.}
    \label{fig:pipeline}
\end{figure}

Our pipeline has three stages: information extraction, inference validation, and temporal analysis. Each of these stages is dedicated to one specific aspect of the extraction process. Fig. \ref{fig:pipeline} pictures the pipeline, numbering its steps \circled{1} through \circled{11} to aid identification in the paragraphs below. The output of the pipeline is a KG separated into two tiers. The first encodes the content of the text in the form of entities and relationships; the second conveys temporal information regarding the relationships themselves. Each step of the pipeline is realized using few-shot prompting (see Appendix \ref{app:prompts} for prompt templates). Our pipeline is fully automated, model-agnostic, and suitable for black-box LLMs, as it does not require accessing the model parameters or fine-tuning.

\subsection{Information Extraction}

The first stage deals with extracting the information implied by the text, both explicitly and implicitly. The model is instructed to infer as much implicit information as possible to improve coverage.

\begin{figure}[ht]
    \centering
    \includegraphics[width=\linewidth]{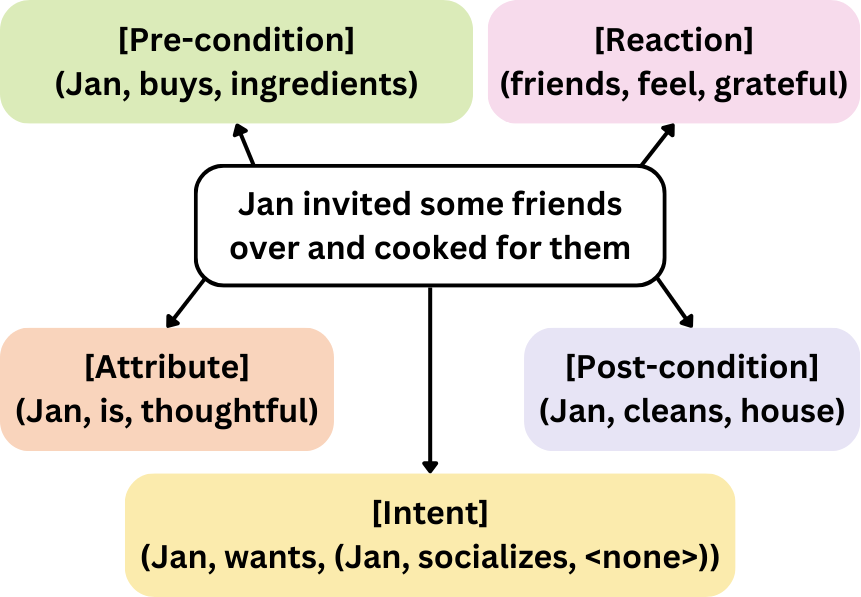}
    \caption{Implicit triplets with various inference types regarding a context sentence.}
    \label{fig:types_ex}
\end{figure}

We start by extracting the entities mentioned in the context sentence \circled{1}, which will constitute the nodes of the KG. The model is also prompted to tag them with their entity type, assigned according to the taxonomy proposed by \citet{COfEE}. We adopted this ontology because of its comprehensiveness as well as its focus on aiding event extraction from text. Then, the model is directed to extract relational triplets representing the content of the text \circled{2}. These are shaped as \trip triplets, where subject and object are entities in the KG. The model first extracts explicit relationships, also reporting the snippet of the context corresponding to the information in each triplet. This step closely aligns with Open Information Extraction (OIE), an NLP task that can be performed at a high level by LLMs \cite{OIEsurvey}. Subsequently, the model is asked to extract implicit triplets, namely those that express implicit meaning. This task can be seen as a combination of OIE with inferential and commonsense reasoning, as these triplets should be deduced from the text. 
\rev{At this stage, we do not provide the model with the explicit triplets extracted in the previous step, so that implicit extraction remains conditioned only on the original sentence (\textit{i.e.}, what the model can infer from the text itself), rather than being influenced by a potentially noisy intermediate representation.}
The model is also given guidance in the form of a list of possible relations between the inferred triplets and the information expressed in the text to facilitate the task. 
These relations (referred to as inference types) are adapted from the \textsc{Atomic} taxonomy \cite{Atomic}. Fig. \ref{fig:types_ex} provides an example of each.
A given triplet can report the \textit{pre-conditions} and \textit{post-conditions} of the events in the text or information about the mentioned actors, such as their \textit{intents}, their \textit{emotional reactions}, and the \textit{attributes} they are perceived to have.
For implicit triplets, the subject and object may not belong to the extracted entities and therefore are added to the entity list.

The main challenges of information extraction are related to the well-defined form of the triplets, as not all information mentioned in the context sentences can be expressed in \trip form. An exemplary use case is that of unary relations, \textit{i.e.}, sentences without an object. In this occurrence, the triplet will feature a \textsc{<none>} tag in place of the object. For instance, “Perry was running” is encoded as \textsc{(Perry, wasRunning, <none>)}.

However, representing other grammatical constructs is not as straightforward. One such case is \textit{subordination}, where a subordinate clause acts as the complement of an independent clause. An example is “Jordan heard Bob was looking for her”. Another case is that of \textit{aspect}, namely the expression of an action’s progression in time, which can be realized by prefixing the action verb with an aspectual verb. For instance, in “Chris stopped talking”, the action of talking can be considered the object of the aspectual verb “stopped”.
These occurrences are handled using an approach inspired by RDF reification \cite{RDFreif}, a mechanism that creates new KG nodes representing complete triplets to express statements about them. Similarly, in our method, triplets may include another triplet as their object \circled{3}: this creates a (potentially recursive) nesting structure that fits subordination and aspect. According to this framework, “Jordan heard Bob was looking for her” is encoded as \textsc{(Jordan, heard, (Bob, wasLookingFor, Jordan))}, with \textsc{(Bob, wasLookingFor, Jordan)} being added to the nodes of the KG. The same method is applied for representing sentences that use aspectual verbs. In this case, the general action being described becomes the nested triplet, while the verb describing its progress is used as the relationship. Compared to the use of reification for subordination, some additional adjustments are required to improve clarity, such as repeating the subject. Recalling the example above, the representation for “Chris stopped talking” is \textsc{(Chris, stopped, (Chris, isTalking, <none>))}.

The reification of triplets representing subordinate clauses poses a new problem, \textit{i.e.}, depending on the semantics of the respective independent clauses, the events and states described by the nested triplets can have different truth values, as implied by their modality. Modality is a feature that \textit{expresses the speaker’s degree of commitment to the events being referred to in a text} \cite{ModalityTimeML}, and as such it significantly influences the meaning that can be attributed to a nested triplet. Possible modalities include \textit{factive} and \textit{asserted}, concerning events implied to be true or \textit{referred to as if [they] really occurred} \cite{COfEE}, as opposed to \textit{modal}, \textit{conditional}, and even \textit{negative}. Nested triplets whose modality is \textit{asserted}, \textit{factive}, or \textit{negative} can be directly included in the KG. However, events with different modalities are uncertain, hence they cannot be added to the KG as-is. Ideally, the LLM should identify the modality of each nested triplet. However, initial experiments showed various models had difficulties in tagging triplets, achieving an accuracy of about 65\% when asked to identify whether a relational triplet was \textit{asserted}. As a result, we decided not to consider nested triplets as valid by themselves. This decision was also supported by the observation that triplets whose truth value is fairly clear should be extracted by the model as implicit relationships.

Lastly, the model is given additional instructions to improve the formal quality and homogeneity of the extracted triplets. The LLM is instructed to use the present tense for each triplet’s relationship. Furthermore, the relationship should be a single verb, with exceptions for prepositions in phrasal verbs and modal verbs prefixing an action verb. Concerning the triplets’ objects, indirect complements should be reported by including their preposition in the relationship and using the relevant KG entity as the object, \textit{e.g.}, “The bracelet belonged to Jill” is encoded as \textsc{(bracelet, belongsTo, Jill)}. Additionally, subject complements should be reported in the object for clarity, \textit{e.g.}, “Todd is athletic” becomes \textsc{(Todd, is, athletic)}.

\subsection{Inference Validation}

The second stage examines implicit triplets, filtering out and correcting potentially unrealistic or unsubstantiated inferences to improve the precision of the KG. This is complementary to the previous stage which focuses on coverage over accuracy.

In the inference challenge step \circled{5}, inferred triplets are individually analyzed by the model to determine whether they are unsupported and should therefore be discarded. At this stage, the model acts as its own critic \cite{KnowledgeDistil}. For every discarded triplet, the model must also provide a justification. Since reliable confidence estimation for black-box LLMs remains challenging \cite{confEstSurvey, selfConsistency}, we guide this validation step through prompt instructions and examples of deducible and non-deducible triplets, implicitly tuning the model's discard criterion.

When a triplet is discarded, it undergoes inference correction \circled{6}. In this step, the model is given the triplet along with the reason why it was discarded (generated in the previous step) and it is asked to issue a correction. The LLM is required to strike a balance between addressing the reason for which the original triplet was discarded and not completely subverting its meaning. In case this is impossible (\textit{e.g.}, if the original triplet’s meaning would need to be significantly altered), the prompt instructs the model to simply avoid correcting the triplet. Note that the statement generated as a correction is not automatically accepted, as it also undergoes inference challenge and, potentially, correction. A simple safeguard is implemented to prevent triplets from entering an infinite challenge-discard-correction loop, \textit{i.e.}, after a triplet (and its subsequent corrections) has been discarded three times, it is rejected without being corrected again.

Two additional steps are implemented to complete the inference validation stage. Duplicate removal \circled{4} is applied to every triplet before inference challenge and allows to directly discard it in case it is redundant, \textit{i.e.}, semantically equivalent to a statement in the set of explicit triplets and validated implicit triplets. The redundancy check is performed by the model itself via few-shot prompting. Inference explanation \circled{7}, instead, aims to shed light on the model’s inference process. For every validated implicit triplet, the LLM is instructed to provide a set of explicit triplets that explain its inference. While these premises cannot be considered a rigorous explanation of the process that generated a specific triplet, they can still offer some insight into how the model deduces implicit triplets.

\subsection{Temporal Analysis}

The last stage focuses on the temporal information regarding the extracted triplets, an aspect that was not considered up to this point. In the information extraction stage, temporal metadata was purposely ignored, making the triplets more formally homogeneous. However, temporal relations become relevant when considering the set of triplets stemming from a given context. For this reason, this stage focuses on the triplets’ position in time, both in absolute terms and relative to each other.

The statements are studied individually to determine whether they are events (defined as situations that happen or occur, both punctual and extended in time) or states (defined as conditions or circumstances that hold true) \circled{8}. At the same time, the model is asked to extract, for all triplets that have it, an absolute temporal reference that grounds the event to a specific moment in time or a duration \circled{9}.

After this process is complete, the model is asked to examine the pairwise temporal relationships between triplets \circled{10} to generate a timeline regarding the events described in the context sentence. The model assigns a relation type to each pair of triplets, according to the following taxonomy: \textit{before}, \textit{after}, \textit{while}, or \textit{none} \cite{EventPlus}. As a safeguard against hallucinations and output variability, triplet pairs are submitted in both possible orders. After parsing the model output, the specified relationship is only assigned if the two pairs are tagged with consistent types \circled{11}, otherwise, the triplets are considered unrelated. Note that the only consistent relationship types for the two permutations of a triplet pair are \textit{before}-\textit{after} and \textit{while}-\textit{while}.


\section{Experimental Setup}
We test our methodology on two target LLMs: Mistral Large 2 \cite{Mistral} and GPT-4o mini \cite{GPT4o}. \rev{We run experiments by executing the pipeline on 30 English-language context sentences in total, 15 randomly drawn from the SocialIQA dataset \cite{SocialIQA} and 15 from premises in the COPA dataset \cite{copa}.}

\rev{Human judgments were collected from two groups. SocialIQA was annotated by 206 volunteer university students of a variety of nationalities. Their median age was 24, and the gender split was 76.6\% male and 23.4\% female. COPA was annotated by 101 MTurk workers. They were compensated \$3.50-\$4.50 per assignment, proportional to form length, and were recruited under strict quality constraints (at least 100 approved HITs and an approval rate of at least 98\%). We did not collect additional self-reported demographics beyond these platform qualifications. To ensure annotation reliability, each assignment included five attention-check questions based on obviously false fabricated triplets. We excluded annotators who failed at least one check, retaining 205 university students and 75 MTurk annotators for analysis.}

\rev{Data was collected anonymously via Microsoft Forms} and free-text fields were screened to avoid releasing personal data. The form has five sections that mirror the stages of our pipeline: \textit{triplet classification} \circledLett{a}, \textit{inference correction review} \circledLett{b}, \textit{event/state classification} \circledLett{c}, \textit{timing comparison} \circledLett{d}, and \textit{model error correction} \circledLett{e}. For task manageability, we split it into six forms (three per model), each covering five context sentences, and each respondent completed one form. In \circledLett{a}, participants label each triplet as factual, deducible, or wrong. In \circledLett{b}, they rate the model’s discard decisions and, if they agree with the discard, the model’s stated reasons and corrections for a randomly selected discarded triplet per sentence. In \circledLett{c}, they label triplets as events or states, and in \circledLett{d}, they judge temporal relations. Finally, in \circledLett{e}, they review the final triplet set, flag any removals, and add missing triplets.

\section{Results and Discussion}


The main metric for comparison is Model-Human Agreement (MHA), \textit{i.e.}, the ratio of questions for which human consensus agrees with the model answer. For closed questions featured in \textit{triplet classification}, \textit{event/state classification}, and \textit{timing comparison}, the human consensus outcome is computed by majority vote and can be directly compared to the model answer. In these cases, direct comparison also allows us to compute Cohen’s $\kappa$ statistic to measure inter-rater reliability between models and humans. Questions in the \textit{inference correction review} section, instead, directly ask the evaluator about their degree of agreement with model decisions. In this case, the human consensus is measured in terms of polarity, \textit{i.e.}, each answer is assigned an agreement score (1 for \textit{fully agree}, 0 for \textit{somewhat agree}, -1 for \textit{disagree}). The evaluators are collectively considered to concur with the model decision if the average of these scores is positive. Lastly, considering triplet removal in \textit{model error correction}, we deem humans to be in agreement with a model-generated triplet if less than 50\% of them flag it as wrong.
On the other hand, questions in \textit{model error correction} that ask humans to add missing triplets are inspected to reveal their inference type, their relationship with the model output, and whether they are equivalent to triplets the model previously generated and discarded. \rev{This analysis helps localize coverage gaps: \textit{Fact} additions indicate missing explicit extraction; non-\textit{Fact} additions indicate missing implicit extraction; overlaps with discarded triplets point to over-pruning in validation; and modification-type additions reflect extraction errors in existing triplets.}
We also compare how the outputs of the two LLMs overlap on the same sentences. Each triplet from one model is checked against the triplets generated by the other model and subsequently classified as either an \textit{exact match} (if identical), a \textit{semantic match} (if semantically equivalent), or \textit{no match}.

\begin{table}[t]
\centering
\resizebox{\columnwidth}{!}{%
\begin{tabular}{>{\raggedright\arraybackslash}l
                >{\raggedright\arraybackslash}l
                cc
                cc}
\hline
 &  &
\multicolumn{2}{c}{\textbf{SocialIQA}} &
\multicolumn{2}{c}{\rev{\textbf{COPA}}} \\
\textbf{Section} & \textbf{Metric} &
\textbf{Mistral} & \textbf{GPT} &
\rev{\textbf{Mistral}} & \rev{\textbf{GPT}} \\
\hline
Triplet & MHA & 66.2\% & 55.9\% & 74.9\% & 64.9\%\\
classification & $\kappa$ & 0.46 & 0.25 & 0.44 & 0.43\\
\rowcolor{black!10}Event/state & MHA & 94.7\% & 86.3\% & 87.7\% & 90.9\%\\
\rowcolor{black!10}classification & $\kappa$ & 0.89 & 0.72 & 0.73 & 0.82\\
Timing & MHA & 60.0\% & 33.3\% & 45.5\% & 33.3\%\\
comparison & $\kappa$ & 0.46 & 0.15 & 0.19 & 0.19\\
\rowcolor{black!10}ICR: discard & MHA & 75.0\% & 50.0\% & 100\% & 88.9\%\\
ICR: reason & MHA & 100\% & 80.0\% & 100\% & 88.9\%\\
\rowcolor{black!10}ICR: correction & MHA & 83.3\% & 100\% & 100\% & 88.9\%\\
MEC: removal & MHA & 96.7\% & 95.2\% & 85.9\% & 94.4\%\\
\hline
\end{tabular}%
}
\caption{LLM-human agreement metrics by section; see Section~5 for metric definitions. In ICR, \textit{discard} is computed on reviewed discarded triplets, while \textit{reason} and \textit{correction} are conditional follow-up scores computed only when annotators reached those questions.}
\label{table:summary}
\end{table}

In general, Model-Human Agreement is moderate and varies across evaluation sections, with Mistral outperforming GPT in most settings (Table~\ref{table:summary}). Both models attain their highest MHA in \textit{model error correction}, followed by \textit{event/state classification}, while \textit{timing comparison} yields the lowest agreement metrics. According to \citeposs{CohenK} interpretation, $\kappa$ indicates weak-to-minimal agreement for \textit{triplet classification} and \textit{timing comparison}, whereas agreement is substantially higher for \textit{event/state classification} (Table~\ref{table:summary}).

At the output level, GPT is consistently more concise than Mistral (Table~\ref{table:output}). Nonetheless, neither model achieves full coverage, as reflected by substantial human additions (Table~\ref{table:output}). Hallucinated implicit relationships are rare. \rev{They appear pre-validation for Mistral on SocialIQA but are removed during validation, and are absent on COPA for both models (Table~\ref{table:output}). Lastly, model outputs overlap to some extent. Overlap is higher on SocialIQA, while on COPA it decreases, especially for Mistral, consistent with Mistral producing a larger volume of triplets and thus accumulating more unmatched relationships (Table~\ref{table:model-overlap}).}

\begin{table}[t]
\centering
\resizebox{\columnwidth}{!}{%
\begin{tabular}{>{\raggedright\arraybackslash}m{0.50\columnwidth}cccc}
\hline
 & \multicolumn{2}{c}{\textbf{SocialIQA}} & \multicolumn{2}{c}{\rev{\textbf{COPA}}} \\
\textbf{Metric} & \textbf{Mistral} & \textbf{GPT} & \rev{\textbf{Mistral}} & \rev{\textbf{GPT}} \\
\hline
Generated triplets (median) & 9 & 6 & 12 & 5\\
\rowcolor{black!10}Human additions (median) & 13 & 16 & 16 & 6\\
\begin{tabular}[c]{@{}l@{}}
Human additions overlap\\
with discarded triplets
\end{tabular}
& 9.6\% & 12.5\% & 1.6\% & 5.9\%\\
\rowcolor{black!10}Human removal rate & 3.3\% & 4.8\% & 14.1\% & 5.8\%\\
Hallucinated triplets (pre-validation) & 8.0\% & 0\% & 0\% & 0\%\\
\rowcolor{black!10}Hallucinated triplets (post-validation) & 0\% & 0\% & 0\% & 0\%\\
\hline
\end{tabular}%
}
\caption{Summary statistics for the outputs of the target models, averaged over the context sentences.}
\label{table:output}
\end{table}

\begin{table}[t]
\centering
\resizebox{\columnwidth}{!}{%
\begin{tabular}{>{\raggedright\arraybackslash}l c c c c}
\hline
 & \multicolumn{2}{c}{\textbf{SocialIQA}} & \multicolumn{2}{c}{\rev{\textbf{COPA}}}\\
\textbf{Match type} & \textbf{Mistral} & \textbf{GPT} & \rev{\textbf{Mistral}} & \rev{\textbf{GPT}}\\
\hline
Exact match & 23.9\% & 32.3\% & 10.3\% & 27.5\%\\
\rowcolor{black!10}Semantic match & 20.9\% & 33.3\% & 13.6\% & 27.5\%\\
No match & 55.2\% & 34.4\% & 76.1\% & 44.9\%\\
\hline
\end{tabular}%
}
\caption{Overlaps in the model outputs by match type.}
\label{table:model-overlap}
\end{table}

\begin{figure}[t]
    \centering
    \includegraphics[width=\linewidth]{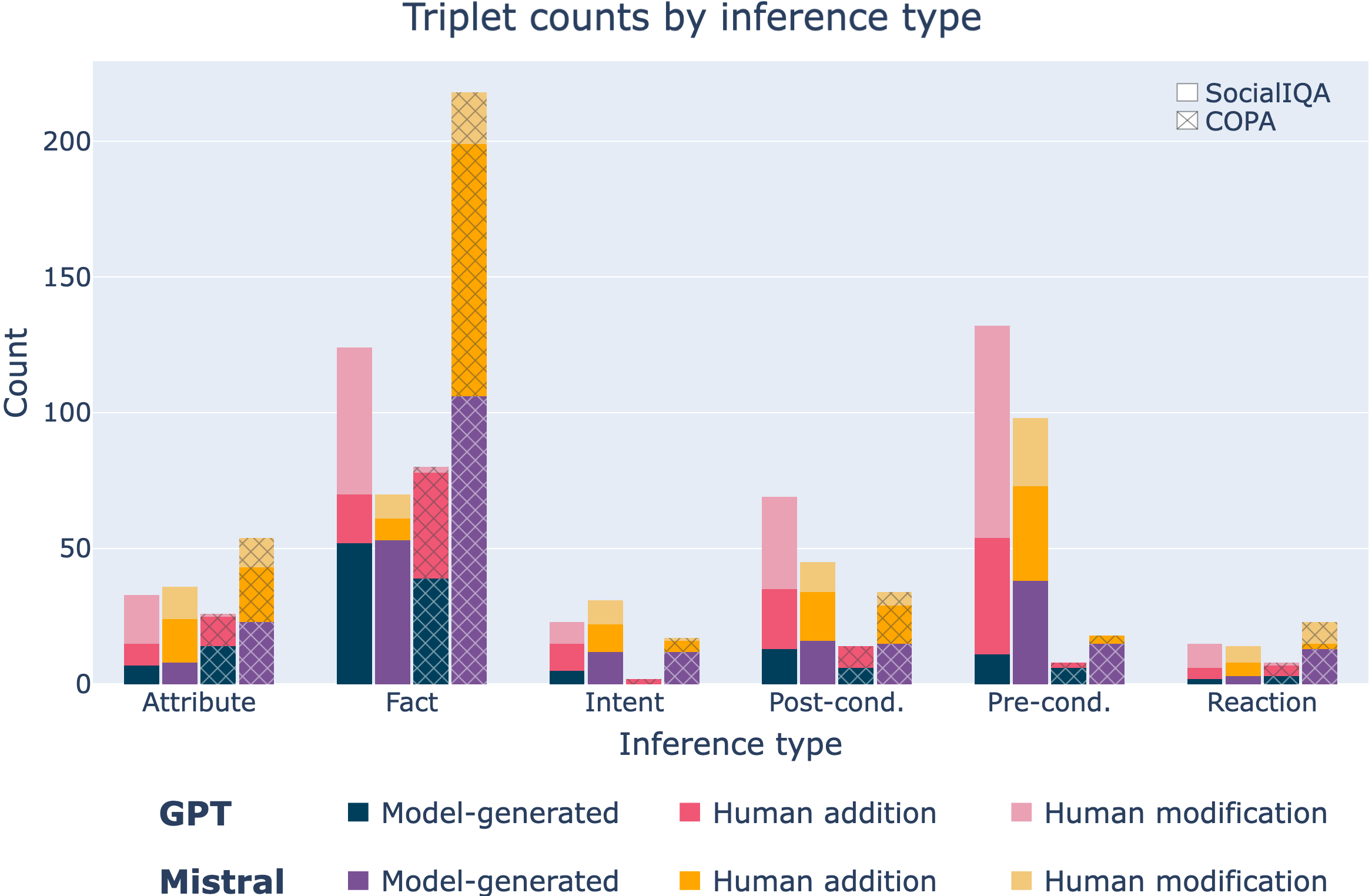}
    \caption{Triplet counts grouped by inference type. Each stack comprises LLM-generated triplets, human additions with no correspondence to the output, and human additions that modify an LLM-generated triplet.}
    \label{fig:inf_types}
\end{figure}

\paragraph{RQ1: Inference Drivers} A noticeable difference between humans and LLMs concerns the interpretation of explicit (or \textit{factual}) and implicit (or \textit{deducible}) triplets. In particular, many triplets the models extracted as \textit{deducible} were instead deemed \textit{factual} by the evaluators (see Table \ref{table:conf_full}). \rev{This pattern holds across both datasets} and suggests both models adopt a fairly literal interpretation of explicit triplets, requiring them to be mentioned in the text almost verbatim. On the other hand, humans tend to classify triplets that express the content of the text using different wording as \textit{factual}. Additionally, examining the inference types of human and model-generated triplets reveals further patterns. \rev{Fig. \ref{fig:inf_types} highlights a dataset shift where COPA is more \textit{fact}-centric, with comparatively fewer \textit{intent} (and, more generally, fewer non-fact) inferences than SocialIQA.} Evaluators also frequently propose \textit{pre-conditions}, \textit{post-conditions}, and \textit{attributes} missing from model outputs, showing that both models under-generate these inference types relative to human additions and modifications (Fig. \ref{fig:inf_types}). \rev{This gap is especially visible for GPT on SocialIQA, while on COPA the mismatch concentrates more on the non-factual inference types, consistent with the dataset’s shorter, more fact-oriented contexts.}

\begin{table}[t]
\centering
\resizebox{\columnwidth}{!}{%
\begin{tabular}{@{}c@{}}

\begin{tabular}{c|ccc @{\hspace{1.4em}} ccc}
 & \multicolumn{3}{c}{\textbf{SocialIQA (Mistral)}} & \multicolumn{3}{c}{\rev{\textbf{COPA (Mistral)}}}\\
 & $\ell_h=\mathbf f$ & $\ell_h=\mathbf d$ & $\ell_h=\mathbf w$
 & \rev{$\ell_h=\mathbf f$} & \rev{$\ell_h=\mathbf d$} & \rev{$\ell_h=\mathbf w$}\\
\hline
$\ell_m=\mathbf f$ &
\cellcolor{LimeGreen!25}33.8\% & 2.8\% & 0.0\% &
\cellcolor{LimeGreen!25}12.3\% & 3.0\% & 1.0\% \\
$\ell_m=\mathbf d$ &
9.9\% & \cellcolor{LimeGreen!25}28.2\% & 2.8\% &
2.5\% & \cellcolor{LimeGreen!25}59.6\% & 6.9\% \\
$\ell_m=\mathbf w$ &
0.0\% & 18.3\% & \cellcolor{LimeGreen!25}4.2\% &
0.0\% & 11.8\% & \cellcolor{LimeGreen!25}3.0\% \\
\end{tabular}

\\[35pt]

\begin{tabular}{c|ccc @{\hspace{1.4em}} ccc}
 & \multicolumn{3}{c}{\textbf{SocialIQA (GPT)}} & \multicolumn{3}{c}{\rev{\textbf{COPA (GPT)}}}\\
 & $\ell_h=\mathbf f$ & $\ell_h=\mathbf d$ & $\ell_h=\mathbf w$
 & \rev{$\ell_h=\mathbf f$} & \rev{$\ell_h=\mathbf d$} & \rev{$\ell_h=\mathbf w$}\\
\hline
$\ell_m=\mathbf f$ &
\cellcolor{LimeGreen!25}36.8\% & 1.5\% & 1.5\% &
\cellcolor{LimeGreen!25}29.7\% & 9.5\% & 2.7\% \\
$\ell_m=\mathbf d$ &
25.0\% & \cellcolor{LimeGreen!25}19.1\% & 0.0\% &
10.8\% & \cellcolor{LimeGreen!25}28.4\% & 5.4\% \\
$\ell_m=\mathbf w$ &
2.9\% & 13.2\% & \cellcolor{LimeGreen!25}0.0\% &
1.4\% & 5.4\% & \cellcolor{LimeGreen!25}6.8\% \\
\end{tabular}

\end{tabular}%
}
\caption{Confusion matrices for LLM-generated triplets from the triplet-classification section, where $\ell_h$ is the human consensus label, $\ell_m$ is the model-assigned label, $\mathbf f$ stands for \textit{factual}, $\mathbf d$ stands for \textit{deducible}, and $\mathbf w$ stands for \textit{wrong}.}
\label{table:conf_full}
\end{table}

\begin{table}[t]
\centering
\resizebox{\columnwidth}{!}{%
\begin{tabular}{>{\raggedright\arraybackslash}l c c}
\hline
\textbf{Section} & \textbf{SocialIQA} & \rev{\textbf{COPA}}\\
\hline
Triplet classification & 83.2\% & 79.9\%\\
\rowcolor{black!10}Inference correction review & 65.1\% & 77.6\%\\
Event/state classification & 81.6\% & 89.3\%\\
\rowcolor{black!10}Timing comparison & 68.5\% & 77.0\%\\
Model error correction & 88.7\% & 84.3\%\\
\hline
\end{tabular}%
}
\caption{Human majority agreement rates by section.}
\label{table:cons}
\end{table}

\paragraph{RQ2: Strictness}
\rev{We operationalize \textit{strictness} as the tendency to label candidate inferences as \textit{wrong} as defined in Table~\ref{table:conf_full}. On SocialIQA, both models prove much stricter than human evaluators. The model-induced label split differs significantly from the human distribution (chi-squared test; p-val<0.001 for both models). This is also supported by humans discarding significantly fewer triplets than the models (one-sided binomial test; p-val<0.001 for both models). This gap is directly visible in Table~\ref{table:conf_full}. For GPT on SocialIQA, humans label only $1.5\%$ of triplets as \textit{wrong}, while for Mistral, the human \textit{wrong} rate is $7.0\%$. In practice, many triplets rejected as \textit{wrong} by the models are instead judged \textit{factual} or \textit{deducible} by the crowd, indicating an over-pruning tendency.}

\rev{On COPA, the strictness picture changes and the label distributions become more aligned. Under the same chi-squared test, we do not reject compatibility between model and human label distributions (Mistral p-val=0.1346, GPT p-val=0.9407; n.s.). Likewise, the strictness-rate comparison does not provide evidence that humans discard fewer triplets than the models (one-sided binomial test; Mistral p-val=0.0647, GPT p-val=0.7055; n.s.). A key driver of this shift is a substantial increase in human strictness. The human \textit{wrong} rate rises from $7.0\%$ to $10.9\%$ for Mistral and, most notably, from $1.5\%$ to $14.9\%$ for GPT (Table~\ref{table:conf_full}). A plausible explanation is that COPA contexts are shorter and more fact-oriented, providing less contextual support for speculative inferred triplets. As a result, evaluators more often deem such inferences unsupported and label them as \textit{wrong}. Conversely, SocialIQA describes richer social situations where multiple implicit details can be reasonably inferred, leading humans to retain many inferences that models discard.
This human-side shift is also reflected in inference correction review (ICR). Discard decisions are less contested on COPA than on SocialIQA (\textit{e.g.}, GPT disagreement-majority drops from 50.0\% to 11.1\%, with average polarity increasing from 0.456 to 0.684), indicating stronger human alignment with discarding in the more fact-oriented setting.
Consistently, the post-validation model error correction stage shows stricter human filtering on COPA, with higher removal rates (\textit{e.g.}, 14.1\% vs. 3.3\% for Mistral; Table~\ref{table:output}).}

\paragraph{RQ3: Timing Information}
\rev{LLMs perform well in \textit{event/state classification}, with both models reporting high agreement with humans (Table~\ref{table:summary}). By contrast, \textit{timing comparison} remains the most challenging component with the lowest agreement. A recurring source of disagreement is the option “\textit{No clear relation}”. While humans rarely select it, models abstain substantially more often, both on SocialIQA ($33.3\%$ vs.\ $6.7\%$) and on COPA ($34.8\%$ overall; GPT $50.0\%$, Mistral $18.2\%$, vs.\ $0\%$ for human consensus). This suggests that models are conservative when temporal relations are not explicitly stated, whereas humans more readily infer event order from implicit cues. Qualitative inspection indicates that several missed relations could be recovered from commonsense event sequences or chaining other explicit triplets, highlighting temporal structure recovery as a key bottleneck.}

\paragraph{RQ4: Human Consensus}
\rev{Human consensus is consistently above random chance across all sections and both datasets (one-sided t-tests vs. chance; all p-val<0.001), indicating that crowd judgments provide a reliable signal. At the same time, consensus strength varies by section (Table~\ref{table:cons}). On SocialIQA, majority agreement is highest in \textit{triplet classification} (83.2\%), \textit{event/state classification} (81.6\%), and \textit{model error correction} (88.7\%), while it drops in \textit{inference correction review} (65.1\%) and \textit{timing comparison} (68.5\%). On COPA, we observe similarly strong agreement overall, with the lowest consensus again occurring in \textit{inference correction review} and \textit{timing comparison}. These patterns suggest that disagreement is not random but concentrated in the most subjective steps of the pipeline (\textit{e.g.}, judging whether an inference is sufficiently supported and recovering implicit temporal structure) where multiple interpretations may remain plausible even for careful annotators.}

\section{Conclusion}
We introduced the task of implicit information extraction (IIE) together with a methodology for studying it with LLMs using few-shot prompting to build a structured knowledge graph. The pipeline extracts relational triplets, validates and corrects implicit inferences while explicitly stating their premises, and analyzes temporal relations among validated triplets. \rev{Our results show moderate model--human agreement with some variation across sub-tasks and datasets. Models are more conservative than humans on SocialIQA, while on COPA the gap attenuates as humans adopt stricter rejection criteria in fact-oriented contexts.} Both models also show coverage limitations and distinct behaviors across sub-tasks, suggesting that the methodology could benefit from tuning to the target LLM and domain.

\section*{Limitations}

One potential limitation of this study is that human judgments of implicit meaning are inherently variable, since text interpretation can depend on annotator background and reading strategies. We address this by collecting multiple independent judgments from a large annotator pool (300+ participants) and by filtering low-quality responses via attention checks. However, the evaluation still covers a limited number of examples per dataset and a small set of LLMs, and may therefore not capture the full variability of implicit interpretation. Additionally, cross-dataset comparisons should be interpreted with caution, since the study was extended to include a second dataset at a later stage and the annotation protocol therefore did not fully control for annotator-population differences across datasets. As a result, some observed differences may reflect annotator-population effects in addition to dataset properties.

\section*{Ethical Considerations}

The lack of transparency about how Large Language Models process and interpret text poses serious ethical questions, particularly considering their widespread adoption in professional fields such as scientific research, journalism, and law, as well as their popularity among the general public for personal use. Our work aims to shed light on how LLMs interpret the implicit meanings embedded in text, thus deepening our understanding of these models and contributing to their interpretability. We believe this will be beneficial in enabling a more trustworthy adoption of Large Language Models.

\bibliography{custom}

\appendix


\revblock{
\section{Comparison with Natural Language Inference (NLI)}
\label{app:nli-sanity}

This appendix section reports a set of auxiliary Natural Language Inference (NLI) experiments used to compare a NLI baseline with the human judgments and the pipeline's internal \emph{discard} decisions.
The goal is to quantify how a strong off-the-shelf NLI model behaves when asked to validate extracted triplets against the original context, how it compares with LLM and human judgments and to what extent it can be used as an external validation tool.

\paragraph{Experimental Setup.}
For each triplet, we build an NLI pair which consists of
(i) a \textit{premise}, which is the original context sentence, and
(ii) a \textit{hypothesis}, which is a verbalized version of the extracted triplet. This is because NLI expects text, not structured triplets.
As NLI model, we use \texttt{microsoft/deberta-large-mnli}\footnote{https://huggingface.co/microsoft/deberta-large-mnli} and record the softmax probabilities
$p_{\text{ent}}$ (entailment), $p_{\text{neu}}$ (neutral), and $p_{\text{con}}$ (contradiction).
We also report \textit{entail\_rate} as the fraction of items where entailment is the argmax label.

We investigate and report three complementary checks:
\textbf{(Check~1)} NLI probabilities grouped by the human label (\textsc{Factual}/\textsc{Deducible}/\textsc{Wrong}).
\textbf{(Check~2)} grouped by the pipeline (\textit{i.e.}, LLM-side) label (\textsc{Factual}/\textsc{Deducible}/\textsc{Wrong}), where
\textsc{Wrong} corresponds to the model discarding the triplet.
\textbf{(Check~3)} targeted disagreement classification focusing on:
Group A = human non-\textsc{Wrong} but model discarded, and
Group B = human \textsc{Wrong} but model kept.

\subsection{Triplet verbalization for NLI}
\label{app:nli-verbalization}
Since the NLI model consumes textual hypotheses, each triplet must be converted into a short sentence.
We use a lightweight verbalizer with a small inventory of templates and nested triplets are rendered recursively.
Table~\ref{tab:verbalization-rules} summarizes the normalization and templating rules used to map entities/predicates and handle missing arguments.
The verbalization is intentionally lightweight and prioritizes determinism over perfect fluency, so as to avoid introducing additional variability or biases that could come along by having an LLM perform the verbalization step.

\begin{table*}[t]
\centering
\scriptsize
\setlength{\tabcolsep}{3pt}
\renewcommand{\arraystretch}{1.10}
\setlength{\extrarowheight}{0.6pt}
\begin{tabularx}{\linewidth}{@{}%
  >{\raggedright\arraybackslash}p{0.18\linewidth}%
  >{\raggedright\arraybackslash}p{0.26\linewidth}%
  >{\raggedright\arraybackslash}p{0.24\linewidth}%
  >{\raggedright\arraybackslash}X%
@{}}
\toprule
\textbf{Rule} & \textbf{Trigger / input} & \textbf{Deterministic output} & \textbf{Example} \\
\midrule
Token splitting &
entities and predicates in \texttt{camelCase} &
split into spaced tokens &
\makecell[tl]{\texttt{strangeLooks} $\rightarrow$ ``strange looks''\\
\texttt{isUnfamiliar} $\rightarrow$ ``is unfamiliar''} \\

Minimal predicate rewrite &
predicate in a fixed rewrite list &
apply rewrite before verbalization &
\texttt{hasAttribute} $\rightarrow$ ``has'' \\

Missing argument handling &
argument is \texttt{<none>} or missing &
render missing slot as empty and collapse whitespace &
\makecell[tl]{\texttt{(teammates, communicate, <none>)}\\
$\rightarrow$ ``Teammates communicate''} \\

Nested triplets &
object is a triplet &
render inner triplet recursively, attaching via default connector \texttt{that} &
\makecell[tl]{\texttt{(husband, questions,}\\
\texttt{\hspace*{1.0em}(wife, loves, husband))}\\
$\rightarrow$ ``Husband questions that
wife loves husband''} \\

Same-subject connector &
object is a triplet, inner subject equals outer subject, predicate is in fixed list &
\makecell[tl]{drop inner subject and use\\
a predicate-specific connector\\
(\textit{e.g.}, \texttt{to}, \texttt{by}, \texttt{when})} &
\makecell[tl]{\texttt{(teammates, want,}\\
\texttt{\hspace*{1.0em}(teammates, communicate, <none>))}\\
$\rightarrow$ ``Teammates want to communicate''} \\
\bottomrule
\end{tabularx}
\caption{\rev{Set of verbalization rules for converting triplets into NLI hypotheses.}}
\label{tab:verbalization-rules}
\end{table*}

\subsection{Results}
\label{app:nli-results}

\begin{table*}[t]
\centering
\small
\setlength{\tabcolsep}{4pt}
\begin{tabular}{l l l r r r r}
\toprule
\textbf{Dataset} & \textbf{LLM} & \textbf{Group} & \textbf{mean $p_{\text{ent}}$} & \textbf{mean $p_{\text{neu}}$} & \textbf{mean $p_{\text{con}}$} & \textbf{entail\_rate} \\
\midrule
COPA      & GPT-4o mini      & Factual   & 0.959 & 0.040 & 0.001 & 0.968 \\
COPA      & GPT-4o mini      & Deducible & 0.712 & 0.261 & 0.027 & 0.719 \\
COPA      & GPT-4o mini      & Wrong     & 0.533 & 0.415 & 0.052 & 0.636 \\
COPA      & Mistral Large 2  & Factual   & 0.949 & 0.047 & 0.004 & 0.967 \\
COPA      & Mistral Large 2  & Deducible & 0.414 & 0.536 & 0.049 & 0.388 \\
COPA      & Mistral Large 2  & Wrong     & 0.235 & 0.734 & 0.031 & 0.182 \\
SocialIQA & GPT-4o mini      & Factual   & 0.925 & 0.035 & 0.040 & 0.952 \\
SocialIQA & GPT-4o mini      & Deducible & 0.560 & 0.435 & 0.005 & 0.583 \\
SocialIQA & GPT-4o mini      & Wrong     & 0.740 & 0.239 & 0.021 & 0.857 \\
SocialIQA & Mistral Large 2  & Factual   & 0.830 & 0.111 & 0.058 & 0.846 \\
SocialIQA & Mistral Large 2  & Deducible & 0.317 & 0.568 & 0.115 & 0.167 \\
SocialIQA & Mistral Large 2  & Wrong     & 0.060 & 0.098 & 0.842 & 0.000 \\
\bottomrule
\end{tabular}
\caption{\rev{\textbf{Check 1.} NLI probabilities grouped by human label (\textsc{Factual}/\textsc{Deducible}/\textsc{Wrong}).}}
\label{tab:check1-human}
\end{table*}

\paragraph{Check 1.}
Across both datasets and both LLMs, \textsc{Factual} triplets tend to receive high entailment probability and entail rate,
suggesting that the NLI model behaves consistently with humans on factual statements.
\textsc{Deducible} triplets typically receive lower entailment and higher neutral probability than \textsc{Factual}, which is expected as commonsense ``deducibility’’ often goes beyond strict textual entailment, so NLI appropriately assigns intermediate support.
The \textsc{Wrong} category shows the largest variability across settings, as the NLI model sometimes assigns high entailment even to triplets that humans label as \textsc{Wrong}.
This does not necessarily indicate a problem with the human labels, but rather, it reflects known failure modes of NLI used as a proxy for support, including treating beliefs, reported speech or accusations as objective facts, and over-reliance on lexical overlap.
We analyze these patterns qualitatively in \ref{app:nli-failures}.

\begin{table*}[t]
\centering
\small
\setlength{\tabcolsep}{4pt}
\begin{tabular}{l l l r r r r}
\toprule
\textbf{Dataset} & \textbf{LLM} & \textbf{Pipeline label} & \textbf{mean $p_{\text{ent}}$} & \textbf{mean $p_{\text{neu}}$} & \textbf{mean $p_{\text{con}}$} & \textbf{entail\_rate} \\
\midrule
COPA      & GPT-4o mini      & Factual   & 0.921 & 0.074 & 0.005 & 0.968 \\
COPA      & GPT-4o mini      & Deducible & 0.730 & 0.236 & 0.034 & 0.750 \\
COPA      & GPT-4o mini      & Wrong     & 0.507 & 0.481 & 0.012 & 0.429 \\
COPA      & Mistral Large 2  & Factual   & 0.920 & 0.070 & 0.010 & 0.941 \\
COPA      & Mistral Large 2  & Deducible & 0.396 & 0.557 & 0.047 & 0.366 \\
COPA      & Mistral Large 2  & Wrong     & 0.074 & 0.889 & 0.038 & 0.000 \\
SocialIQA & GPT-4o mini      & Factual   & 0.852 & 0.102 & 0.046 & 0.905 \\
SocialIQA & GPT-4o mini      & Deducible & 0.759 & 0.238 & 0.003 & 0.813 \\
SocialIQA & GPT-4o mini      & Wrong     & 0.431 & 0.560 & 0.009 & 0.333 \\
SocialIQA & Mistral Large 2  & Factual   & 0.759 & 0.134 & 0.107 & 0.733 \\
SocialIQA & Mistral Large 2  & Deducible & 0.349 & 0.648 & 0.003 & 0.167 \\
SocialIQA & Mistral Large 2  & Wrong     & 0.236 & 0.494 & 0.270 & 0.200 \\
\bottomrule
\end{tabular}
\caption{\rev{\textbf{Check 2.} NLI probabilities grouped by the pipeline (LLM-side) label (\textsc{Factual}/\textsc{Deducible}/\textsc{Wrong}).}}
\label{tab:check2-discard}
\end{table*}

\paragraph{Check 2.}
Grouping by the LLM label provides a complementary view to Check~1.
As expected, triplets that the LLM deems as \textsc{Factual} consistently result in higher entailment probability and rate,
\textsc{Deducible} lies in between \textsc{Factual} and \textsc{Wrong},
typically shifting probability toward neutrality.
This is consistent with the intended semantics that inferred triplets are plausible but not necessarily entailed by the context.
Triplets discarded by the LLM (\textit{i.e.}, \textsc{Wrong}) further shift probability toward neutral and/or contradiction, suggesting that the NLI model agrees with most of the pipeline’s discard decisions.
This is especially evident for COPA with Mistral Large 2 where none of the discarded triplets by the LLM were classified as entailed by NLI.

\begin{table*}[t]
\centering
\small
\setlength{\tabcolsep}{4pt}
\begin{tabular}{l l l r r r r}
\toprule
\textbf{Dataset} & \textbf{LLM} & \textbf{Group} & \textbf{mean $p_{\text{ent}}$} & \textbf{mean $p_{\text{neu}}$} & \textbf{mean $p_{\text{con}}$} & \textbf{entail\_rate} \\
\midrule
COPA      & GPT-4o mini      & A & 0.628 & 0.369 & 0.003 & 0.500 \\
COPA      & GPT-4o mini      & B & 0.604 & 0.335 & 0.062 & 0.750 \\
COPA      & Mistral Large 2  & A & 0.082 & 0.913 & 0.005 & 0.000 \\
COPA      & Mistral Large 2  & B & 0.245 & 0.732 & 0.023 & 0.190 \\
SocialIQA & GPT-4o mini      & A & 0.431 & 0.560 & 0.009 & 0.333 \\
SocialIQA & GPT-4o mini      & B & 0.740 & 0.239 & 0.021 & 0.857 \\
SocialIQA & Mistral Large 2  & A & 0.236 & 0.494 & 0.270 & 0.200 \\
SocialIQA & Mistral Large 2  & B & 0.060 & 0.098 & 0.842 & 0.000 \\
\bottomrule
\end{tabular}
\caption{\rev{\textbf{Check 3.} Disagreement analysis in discarded triplets. Group A = human non-\textsc{Wrong} but model discarded. Group B = human \textsc{Wrong} but model kept.}}
\label{tab:check3-triage}
\end{table*}

\paragraph{Check 3.}
Table~\ref{tab:check3-triage} isolates two disagreement regimes:
(A) triplets labeled non-\textsc{Wrong} by humans but discarded by the LLM pipeline, and
(B) triplets labeled \textsc{Wrong} by humans but kept by the pipeline.
Across settings, these cases tend to be inherently difficult for the NLI model. Compared to Checks 1 and 2, they often show
more neutral mass and lower confidence overall, suggesting limited diagnostic value from NLI alone.
Still, Check~3 is useful as a targeted disagreement analysis.
For GPT outputs, Group~B frequently remains NLI-plausible even when humans mark the triplet as \textsc{Wrong},
a pattern commonly driven by lexical overlap or pragmatic plausibility (see \ref{app:nli-failures}).
Conversely, some settings exhibit the opposite behavior, where disagreement cases concentrate clear contradictions.
Overall, the main takeaway is that disagreement triplets are a concentrated source of ambiguity.
NLI does not consistently side with either humans or the pipeline, but it may help identify which mismatches warrant qualitative inspection.

\subsection{Qualitative analysis of common failure modes}
\label{app:nli-failures}
We summarize representative patterns observed in some of the mismatches.
These should be interpreted primarily as possible limitations of using NLI as an external validation tool.

\paragraph{(1) Reported speech and accusations treated as entailment.}
A recurring failure mode is that both the LLM pipeline and the NLI probe often treat a \emph{speech act} or an \emph{accusation} as a factual claim.
In SocialIQA, the context
\emph{``Robin knew that Kai really wanted her to like her outfit so when Kai asked her opinion, Robin said she loved it''}
reports what Robin \emph{said}, not what she truly believed.
Nevertheless, the hypotheses \emph{``Robin loves outfit''} and \emph{``Robin likes outfit''} are considered factual by the LLM and also receive high $p_{\text{ent}}$ from the NLI model,
even though the text does not imply that Robin genuinely loves/likes the outfit (hence the human \textsc{Wrong} label).
Similarly, in COPA the context \emph{``The sales associate accused the girl of stealing''} does not entail that the girl stole, yet the hypothesis \emph{``girl steals''}, which is \textsc{Wrong} according to humans, is labeled factual by the LLM and judged as entailed by NLI.
Overall, these examples show that linguistic commitment (said/accused/believed) is easily conflated with factuality by both NLI and LLMs, warranting cautious use of NLI as a diagnostic tool for IIE.

\paragraph{(2) Temporal non-realization treated as entailment.}
Hypotheses that incorrectly assume an anticipated or otherwise non-factual event is realized can still be labeled as entailed by NLI.
For instance, in the context \emph{``The man anticipated cold weather on his trip''}, NLI assigns high entailment
to hypotheses that assume realization (\textit{e.g.}, \emph{``Man experiences cold weather''}).
This is somewhat similar to the speech/accusation pattern discussed in the paragraph above in that both blur possibility and factuality,
but it is also conceptually distinct as the source of non-entailment here is temporality and uncertainty (\textit{i.e.}, the text describes the possibility of an event in the future but does not imply its realization).

\paragraph{(3) Ill-formed and underspecified triplets with lexical overlap biasing towards high entailment.}
When the LLM-side pipeline produces triplets that are partially ill-formed or underspecified (\textit{e.g.}, missing necessary arguments or awkward predicate-object structures), humans tend to label them as \textsc{Wrong}.
On the other hand, NLI often becomes overly sensitive to shallow lexical cues and yield high entailment, effectively matching on wording overlap even when the sentence is meaningless.
For instance, the verbalization derived from \texttt{(bar, isAt, <none>)} is incomplete (``Bar is at'') yet it receives high entailment.
Similarly, triplets like \texttt{(teammates, interact, blame)} can lead to syntactically degraded hypotheses that nonetheless attract entailment due to overlap with the premise about \emph{teammates} and \emph{blame}.
Overall, these cases warn against a possible self-reinforcing failure mode where upstream extraction can generate low-quality triplets, and NLI may over-reward surface overlap when the hypothesis is not semantically well formed.

\paragraph{(4) Discourse structure misread as contradiction.}
Another recurring failure mode is that the NLI model sometimes mishandles concessive discourse markers (\textit{e.g.}, \emph{even though}),
treating the clause they introduce as if it were being denied rather than merely contrasted with what follows.
A concrete example appears in SocialIQA with the factual triplet \texttt{(Jesse, hasToDo, homework)} extracted from the context
\emph{``Even though she had homework to do that night, Jesse helped Skylar study''}
Humans label \texttt{(Jesse, hasToDo, homework)} as \textsc{Factual} because the sentence explicitly states that Jesse had homework.
However, the presence of the concessive frame (``Even though \dots, Jesse helped \dots'') can confuse NLI, resulting in a high contradiction score,
as if the second clause (helping Skylar) somehow negates the first (having homework to do).
In addition, this example of failure may be exacerbated by a confusion between an obligation ( \emph{having homework to do})
and actually \emph{doing} the homework (a realized event), which could push the model toward the contradiction judgment.

\subsection{Takeaways}
\label{app:nli-takeaways}
(1) Across datasets and LLMs, \textsc{Factual} triplets receive consistently high entailment scores, suggesting that the NLI model tends to agree with humans and LLM on factual triplets.

\noindent (2) \textsc{Deducible} triplets systematically shift probability toward \emph{neutral} relative to \textsc{Factual}, which is expected as many commonsense inferences are plausible but not licensed by strict textual entailment.

\noindent (3) Grouping by the pipeline (LLM-side) label (Check~2) yields the expected ordering (\textsc{Factual} $>$ \textsc{Deducible} $>$ \textsc{Wrong}) in terms of entailment probability and suggests that many discard decisions correlate with lower NLI support. 

\noindent (4) The main instability arises for \textsc{Wrong} triplets. Especially for triplets that were labeled as \textsc{Wrong} by humans, NLI often assigns high entailment under well-known patterns such as speech/accusations treated as facts, temporal non-realization, and shallow lexical matching in ill-formed hypotheses (Section~\ref{app:nli-failures}). This reinforces that NLI is useful here as a diagnostic probe, not as a replacement for human judgment.

\noindent (5) Disagreement cases (Check~3) are largely a concentrated source of ambiguity. Compared to Checks~1 and 2 they often produce lower confidence NLI outputs and do not reliably align with either humans or the pipeline, making them best suited for targeted qualitative inspection rather than aggregate conclusions.

}

\section{Prompts}\label{app:prompts}

\paragraph{Entity Extraction}

\begin{minted}[
frame=single,
framesep=1mm,
linenos=false,
xleftmargin=2pt,
tabsize=2,
fontsize=\scriptsize,
breaklines,
escapeinside=||,
breaksymbolleft=]{text}
Given a piece of text, extract all relevant entities mentioned in it. Ensure each extracted entity is unique and distinct: if two entities with the same name are mentioned, specify additional information that allows them to be identified unambiguously. Otherwise, entity names need to be as succinct as possible: a single noun suffices, so avoid mentioning attributes or other information if it is not necessary for disambiguation purposes.

Additionally, associate a tag to each entity indicating the entity type, choosing among the following: <per>, a distinct person or set of people; <ani>, a distinct animal or set of animals; <org>, an organization characterized by a formally established association (e.g., businesses, sports teams, music groups, government units, educational institutions); <gpe>, a geographical, social, or political entity, including continents, nations, counties, districts, states, provinces, and cities; <fac>, a facility, building or piece of infrastructure, such as a house, factory, gym, museum, office building, parking garage, airport, train station, etc.; <obj>, any man-made object, artifact, or structure that does not fall under the facility category; <occ>, an occupation in the sense of a role, job, position held by one or more people; <tim>, a reference to a time or duration; <num>, a reference to a number, either as an absolute value or with a unit of measure; <msc>, any other entity that does not fall under the categories above.

Be as concise as possible and provide an answer in the form entity1 <type1>; entity2 <type2>; ...

Example 1:
Text: Cameron decided to have a barbecue and gathered her friends together.
Entities: Cameron <per>; barbecue <msc>; friends <per>

Example 2:
Text: Seth has become a huge fan of Queen ever since Anna suggested he listen to them.
Entities: Seth <per>; Queen <org>; Anna <per>

Example 3:
Text: Max parked his car right where Susan usually parks hers.
Entities: Max <per>; Susan <per>; Max’s car <obj>; Susan’s car <obj>

Text: [context sentence]
Entities:
\end{minted}

\paragraph{Relationship Extraction - Explicit}

\begin{minted}[
frame=single,
framesep=1mm,
linenos=false,
xleftmargin=2pt,
tabsize=2,
fontsize=\scriptsize,
breaklines,
escapeinside=||,
breaksymbolleft=]{text}
Given a piece of text and a list of entities mentioned in it, extract relational triplets in the form of (Subject, Relation, Object) from them, based on what is expressed in the text. Paraphrasing is encouraged to include general relations that could be applied to other similar situations, but it should not be used if it would result in an excessive loss of specificity. This should mainly be applied to common situations, and it should be avoided if the given text mentions a highly specific circumstance. The relation of each triplet must be expressed in the present tense, regardless of how it was worded in the text. The triplets must not contain information related to timing, for instance, the starting time of an event or its duration.

The subjects and objects of each triplet should strictly be selected from the list of entities, unless one of the following occurs: if the text conveys information that can be represented as a unary relation without an object, use <none> as the object; if the text mentions an entity having an attribute or a quality, use the attribute in question as the object.

Furthermore, in the following cases a subordinate sentence should be used as the object:
- Events that introduce a reference to a possible world -- e.g., Mary wanted John to buy some wine -> (Mary, wants, (John, buys, wine)).
- Verbs that introduce an entailment (or presupposition) of the argument’s veracity -- e.g., Mary regrets that she didn’t marry John -> (Mary, regrets, (Mary, doesNotMarry, John)).
- Events that introduce a presupposition about the non-veracity of the argument -- e.g., Lina forgot she had to buy flowers -> (Lina, forgets, (Lina, hasToBuy, flowers)).
- Events that fall into the categories of reporting and perception -- e.g., Tim saw Ron crossing the street -> (Tim, sees, (Ron, crosses, street)).
- Reporting and perception events with negative polarity -- e.g., Sue denied she ate the cake -> (Sue, denies, (Sue, eats, cake)).
- Events introduced by negative particles -- e.g., Greg didn’t want to go home -> (Greg, doesNotWant, (Greg, goes, home)).
- Events accompanied by an aspectual verb -- e.g., Jody stopped talking -> (Jody, stops, (Jody, talks, <none>)).

If one of these applies, the subordinate sentence should first be extracted as a relational triplet, maintaining a credible structure with a subject, a relation, and an object by changing the exact wording from the text where needed. This triplet should then be used as the object. Notice that this structure can be applied recursively, and the object triplet itself could have another triplet as its object.

Additionally, for each relational triplet, extract a reference to the part of the text where the relation is mentioned or hinted to. Simply repeat the relevant snippet between ``, copying it from the given text. Select the shortest possible snippet that contains the relation.

Be as concise as possible and provide an answer in the following form: [(sub1, rel1, obj1) `...`; (sub2, rel2, obj2) `...`; ...].

Example 1:
Text: Jesse was pet sitting for Addison, so Jesse came to Addison’s house and walked their dog.
Entities: [Jesse, Addison, house, dog]
Triplets: [(Jesse, petSittingFor, Addison) `Jesse was pet sitting for Addison`; (Jesse, goesTo, house) `Jesse came to Addison’s house`; (Jesse, walks, dog) `walked their dog`; (Addison, owns, dog) `their dog`]

Example 2:
Text: Kai lost Austin’s phone yesterday morning and felt terrible at losing such a valuable item.
Entities: [Kai, Austin, phone]
Triplets: [(Kai, loses, phone) `Kai lost Austin’s phone`, (Austin, owns, phone) `Austin’s phone`, (Kai, feels, terrible) `felt terrible`, (phone, hasAttribute, valuable) `such a valuable item`]

Example 3:
Text: Lewis mentioned he would have liked to be a musician as a kid.
Entities: [Lewis, musician]
Triplets: [(Lewis, mentions, (Lewis, wouldLike, (Lewis, isA, musician))) `Lewis mentioned he would have liked to be a musician`]

Example 4:
Text: Mona just wanted to sleep after a grueling shift at the hospital.
Entities: [Mona, shift, hospital]
Triplets: [(Mona, wants, (Mona, sleeps, <none>)) `Mona just wanted to sleep`, (shift, isAt, hospital) `shift at the hospital`, (shift, hasAttribute, grueling) `grueling shift`]

Text: [context sentence]
Entities: [extracted entities]
Triplets:
\end{minted}

\paragraph{Relationship Extraction - Implicit}

\begin{minted}[
frame=single,
framesep=1mm,
linenos=false,
xleftmargin=2pt,
tabsize=2,
fontsize=\scriptsize,
breaklines,
escapeinside=||,
breaksymbolleft=]{text}
Given a piece of text and a list of entities mentioned in it, list additional facts that can be deduced or inferred from the text. These must be compatible with and derived from the provided text, even though they are not explicitly stated in the text. List as many facts as can possibly be derived; avoiding false positives is not the main concern.

The facts can include, but should not be limited to, the following: the subject’s intent in performing the specified action; the subject’s or the object’s reaction to the event, emotional or otherwise; the events likely to precede or follow the specified action, such as necessary pre-conditions or subsequent effects; the qualities that can be attributed to the subject or the object based on the event.

Represent these facts as relational triplets in the form of (Subject, Relation, Object). The relation of each triplet must be expressed in the present tense, regardless of how it was worded in the text. The triplets must not contain information related to timing, for instance, the starting time of an event or its duration. Avoid unnecessary repetitions; triplets in the list must not be duplicates of other triplets.

If the subject and/or the object of a triplet is an entity present in the provided list, its name has to be reported verbatim. If the text conveys information that can be represented as a unary relation without an object, use <none> as the object. If the text mentions an entity having an attribute or a quality, use the attribute in question as the object. Lastly, the object can also be a nested triplet to allow for more expressiveness, such as in the following cases:
- Events that introduce a reference to a possible world -- e.g., (Mary, wants, (John, buys, wine)).
- Verbs that introduce an entailment (or presupposition) of the argument’s veracity -- e.g., (Mary, regrets, (Mary, doesNotMarry, John)).
- Events that introduce a presupposition about the non-veracity of the argument -- e.g., (Lina, forgets, (Lina, hasToBuy, flowers)).
- Events that fall into the categories of reporting and perception -- e.g., (Tim, sees, (Ron, crosses, street)).
- Reporting and perception events with negative polarity -- e.g., (Sue, denies, (Sue, eats, cake)).
- Events introduced by negative particles -- e.g., (Greg, doesNotWant, (Greg, goes, home)).
- Events accompanied by an aspectual verb -- e.g., (Jody, stops, (Jody, talks, <none>)).

In any other case, it is crucial that the object is a credible entity. Ideally, it should be a single word, but it can include more if needed.

Be as concise as possible and provide an answer in the following form: [(sub1, rel1, obj1); (sub2, rel2, obj2); ...].

Example 1:
Text: Jesse was pet sitting for Addison, so Jesse came to Addison’s house and walked their dog.
Entities: [Jesse, Addison, house, dog]
Triplets: [(Addison, livesIn, house), (Addison, trusts, Jesse), (dog, isAt, house), (dog, wants, (dog, goesOut, <none>))]

Example 2:
Text: Bailey was babysitting a child for the weekend. Bailey took him to church.
Entities: [Bailey, child, weekend, church]
Triplets: [(Bailey, likes, children), (Bailey, is, religious), (Bailey, attends, church), (Bailey, babysits, <none>)]

Example 3:
Text: Bernie was studying, but then he was interrupted by an incoming call.
Entities: [Bernie, call]
Triplets: [(Bernie, caresAbout, studying), (Bernie, stops, (Bernie, studies, <none>)), (Bernie, answers, call)]

Text: [context sentence]
Entities: [extracted entities]
Triplets:
\end{minted}

\paragraph{Inference Challenge}

\begin{minted}[
frame=single,
framesep=1mm,
linenos=false,
xleftmargin=2pt,
tabsize=2,
fontsize=\scriptsize,
breaklines,
escapeinside=||,
breaksymbolleft=]{text}
You are given a piece of text and a relational triplet in the form (Subject, Relation, Object). The triplet was inferred or derived from the given text, but it is not explicitly mentioned in it. Without being excessively strict, specify whether or not the triplet can be reasonably deduced from the text: if not, add a brief explanation of the reason (maximum 20 words).

If the triplet is not precisely indicated or mentioned in the text, that does not automatically disqualify it from being a reasonable deduction or inference; however, the information the text provides should be a solid enough premise for the conclusion that is the inferred triplet. Note that the triplet might not strictly need a premise from the text to be a valid and reasonable inference.

Simply answer “yes” or “no; <explanation>”. Do not produce an explanation if the answer is “yes”. The first word of the answer must be either “yes” or “no”. If present, the explanation must be succinct and strictly related to the given triplet; it must not exceed 20 words. Do not include anything other than the “yes”/“no” and the short explanation.

Example 1:
Text: Jesse was pet sitting for Addison, so Jesse came to Addison’s house and walked their dog.
Inference: (Jesse, takesCareOf, dog)
CanBeInferred: yes

Example 2:
Text: Leah and Fawn have been classmates for 5 years.
Inference: (Leah, isFriendOf, Fawn)
CanBeInferred: no; being classmates does not necessarily imply friendship.

Example 3:
Text: Quinn was having a talk with their significant other. Quinn poured her heart out.
Inference: (Quinn, wants, (significant other, understands, Quinn))
CanBeInferred: yes

Example 4:
Text: Alex called the school immediately to make a complaint with the principal.
Inference: (Alex, expects, (principal, solves, (Alex, has, complaint)))
CanBeInferred: no; Alex speaking to the principal does not mean he expects them to solve his complaint.

Text: [context sentence]
Inference: [implicit triplet to analyze]
CanBeInferred:
\end{minted}

\paragraph{Inference Correction}

\begin{minted}[
frame=single,
framesep=1mm,
linenos=false,
xleftmargin=2pt,
tabsize=2,
fontsize=\scriptsize,
breaklines,
escapeinside=||,
breaksymbolleft=]{text}
You are given a piece of text, a relational triplet in the form (Subject, Relation, Object), and a brief explanation. The triplet was deduced from the given text, but it is not explicitly mentioned in it; however, the inference was deemed wrong for the reasons clarified in the explanation. Considering the explanation provided, apply a correction to the given triplet to eliminate the reason why the original triplet was discarded and ensure that the corrected triplet can actually be inferred from the text. The correction should address the problematic aspects of the original triplet, as highlighted in the explanation, without excessively modifying it. If it is not possible to correct the issues highlighted by the explanation, do not force a correction and simply answer “none”.

Be as concise as possible. If it is possible to apply a correction, provide only the corrected triplet in the form (Subject, Relation, Object); else, simply answer “none”.

Example 1:
Text: Jesse was pet sitting for Addison, so Jesse came to Addison’s house and walked their dog.
Inference: (Addison, owns, house)
Explanation: The text does not imply that Addison is the legal owner of the house.
Correction: (Addison, livesIn, house)

Example 2:
Text: Leah and Fawn have been classmates for 5 years
Inference: (Leah, isClassmateOf, Fawn)
Explanation: Being classmates does not necessarily imply friendship.
Correction: none

Example 3:
Text: Alex called the school immediately to make a complaint with the principal.
Inference: (Alex, expects, (principal, solves, (Alex, has, complaint)))
Explanation: Alex speaking to the principal does not mean he expects them to solve his complaint.
Correction: (Alex, expects, (principal, listensTo, complaint))

Text: [context sentence]
Inference: [implicit triplet to correct]
Explanation: [reason for discarding the triplet]
Correction:
\end{minted}

\paragraph{Inference Explanation}

\begin{minted}[
frame=single,
framesep=1mm,
linenos=false,
xleftmargin=2pt,
tabsize=2,
fontsize=\scriptsize,
breaklines,
escapeinside=||,
breaksymbolleft=]{text}
You are given a piece of text, an inferred relational triplet, and a list of relational triplets explicitly mentioned in the text. The inferred triplet was derived from the given text, but it is not explicitly mentioned in it. All triplets are in the form (Subject, Relation, Object). Specify which of the explicit relationships (if any) can be used as a premise to explain or derive the inferred triplet.

Be as concise as possible and provide an answer in the following form: [(sub1, rel1, obj1); (sub2, rel2, obj2); ...].

Example 1:
Text: Jesse was pet sitting for Addison, so Jesse came to Addison’s house and walked their dog.
Inference: (dog, isAt, house)
Relationships: [(Jesse, petSittingFor, Addison), (Jesse, cameTo, house), (Jesse, walked, dog), (Addison, owns, dog)]
Premise: [(Jesse, petSittingFor, Addison), (Jesse, cameTo, house)]

Example 2:
Text: Quinn was having a talk with their significant other. Quinn poured their heart out.
Inference: (Quinn, wants, (significant other, understands, Quinn))
Relationships: [(Quinn, talksTo, significant other), (Quinn, poursOut, heart), (heart, belongsTo, Quinn)]
Premise: [(Quinn, talksTo, significant other), (Quinn, poursOut, heart)]

Example 3:
Text: Jan bought a cat at the pet store and brought it home with them.
Inference: (cat, needs, (cat, drinks, water))
Relationships: [(Jan, buys, cat), (Jan, bringsTo, home), (cat, isAt, pet store)]
Premise: []

Text: [context sentence]
Inference: [implicit triplet to explain]
Relationships: [extracted explicit relationships]
Premise:
\end{minted}

\paragraph{Duplicate Removal}

\begin{minted}[
frame=single,
framesep=1mm,
linenos=false,
xleftmargin=2pt,
tabsize=2,
fontsize=\scriptsize,
breaklines,
escapeinside=||,
breaksymbolleft=]{text}
You are given a piece of text, a list of relational triplets in the form [(sub1, rel1, obj1), (sub2, rel2, obj2), ...] and a candidate triplet (subject, relation, object). The list contains triplets that represent information expressed in the text. Determine whether or not the candidate triplet is a duplicate of those in the list, that is, if it expresses the same information as a triplet in the list, whether the two use the exact same wording or a semantically equivalent phrasing.

Be as concise as possible. Simply answer “yes” if the candidate triplet is a duplicate, “no” otherwise. No punctuation.

Text: [context sentence]
Triplets: [extracted relationships]
Candidate: [implicit triplet to analyze]
IsDuplicate:
\end{minted}

\paragraph{Event/State Detection and Temporal Grounding}

\begin{minted}[
frame=single,
framesep=1mm,
linenos=false,
xleftmargin=2pt,
tabsize=2,
fontsize=\scriptsize,
breaklines,
escapeinside=||,
breaksymbolleft=]{text}
You are given a piece of text and a list of relational triplets [(sub1, rel1, obj1); (sub2, rel2, obj2); ...]. For each relational triplet, determine whether it represents an event or a state in the given context. An event is a situation that happens or occurs; an event can be punctual, or it can last for a period of time. A state is a condition or a circumstance that holds true; it can be constant throughout the text, or it can change.

Additionally, for each triplet extract an absolute temporal reference if the text provides it. Such a reference anchors the event to a specific moment in time, such as a day, a year, an hour. References relative to other events should be disregarded. Report the temporal reference between ``. Triplets might not have a temporal reference, in which case the answer should simply be `none`.

Be as concise as possible. Provide an answer by repeating the triplets verbatim, in the same order they were given, and associating each of them to either an <event> tag or a <state> tag and a temporal reference (or `none`), in the form [(sub1, rel1, obj1) <...> `...`; (sub2, rel2, obj2) <...> `...`; ...]

Example 1:
Text: Jesse was pet sitting for Addison, so Jesse came to Addison’s house and walked their dog.
Triplets: [(Jesse, petSitsFor, Addison); (Jesse, goesTo, house); (Jesse, walks, dog); (Addison, trusts, Jesse); (Jesse, likes, dogs)]
Tags: [(Jesse, petSitsFor, Addison) <event> `none`; (Jesse, goesTo, house) <event> `none`; (Jesse, walks, dog) <event> `none`; (Addison, trusts, Jesse) <state> `none`; (Jesse, likes, dogs) <state> `none`]

Example 2:
Text: Aubrey tried and got Kendall to go to the dance recital next Wednesday at the town square.
Triplets: [(Aubrey, invites, Kendall); (Kendall, attends, recital); (recital, locatedAt, town square)]
Tags: [(Aubrey, invites, Kendall) <event> `none`; (Kendall, attends, recital) <event> `next Wednesday`; (recital, locatedAt, town square) <state> `none`]

Text: [context sentence]
Triplets: [extracted relationships]
Tags:
\end{minted}

\paragraph{Temporal Relation Extraction}

\begin{minted}[
frame=single,
framesep=1mm,
linenos=false,
xleftmargin=2pt,
tabsize=2,
fontsize=\scriptsize,
breaklines,
escapeinside=||,
breaksymbolleft=]{text}
You are given a piece of text and a list of pairs of relational triplets in the form [((s1, r1, o1), (s2, r2, o2)); ((s3, r3, o3); (s4, r4, o4)); ...]. Based on the text, derive the relative temporal relationship between each pair of triplets by marking it with a tag: <before> if the first triplet of the pair occurs before the second; <after> if the first triplet of the pair occurs after the second; <while> if the first and the second triplet occur simultaneously; <none> if no defined temporal relationship between the two triplets can be derived from the text.

Provide an answer by repeating the pairs of triplets in the same order they were given and associating each of them to their temporal relation tag. Answer in the form [((s1, r1, o1), (s2, r2, o2)) -> ...; ((s3, r3, o3); (s4, r4, o4)) -> ...; ...]. Be as concise and brief as possible and avoid answering with anything other than the triplet pairs and the tags.

Example 1:
Text: Jesse was pet sitting for Addison, so Jesse came to Addison’s house and walked their dog.
Triplet pairs: [((Jesse, petSittingFor, Addison), (Jesse, cameTo, house)), ((Jesse, petSittingFor, Addison), (Jesse, walked, dog)), ((Jesse, cameTo, house), (Jesse, petSittingFor, Addison)), ((Jesse, cameTo, house), (Jesse, walked, dog)), ((Jesse, walked, dog), (Jesse, petSittingFor, Addison)), ((Jesse, walked, dog), (Jesse, cameTo, house))]
Tags: [((Jesse, petSittingFor, Addison), (Jesse, cameTo, house)) -> <while>, ((Jesse, petSittingFor, Addison), (Jesse, walked, dog)) -> <while>, ((Jesse, cameTo, house), (Jesse, petSittingFor, Addison)) -> <while>, ((Jesse, cameTo, house), (Jesse, walked, dog)) -> <before>, ((Jesse, walked, dog), (Jesse, petSittingFor, Addison)) -> <while>, ((Jesse, walked, dog), (Jesse, cameTo, house)) -> <after>]

Example 2:
Text: Cameron decided to have a barbecue and gathered her friends together.
Triplet pairs: [((Cameron, hosted, barbecue), (Cameron, gathered, friends)), ((Cameron, gathered, friends), (Cameron, hosted, barbecue))]
Tags: [((Cameron, hosted, barbecue), (Cameron, gathered, friends)) -> <after>, ((Cameron, gathered, friends), (Cameron, hosted, barbecue)) -> <before>]

Text: [context sentence]
Triplet pairs: [pairs of extracted triplets]
Tags:
\end{minted}

\section{Additional Results}\label{app:results}

\subsubsection*{Inference Correction}

\begin{figure}[ht]
    \centering
    \includegraphics[width=\linewidth]{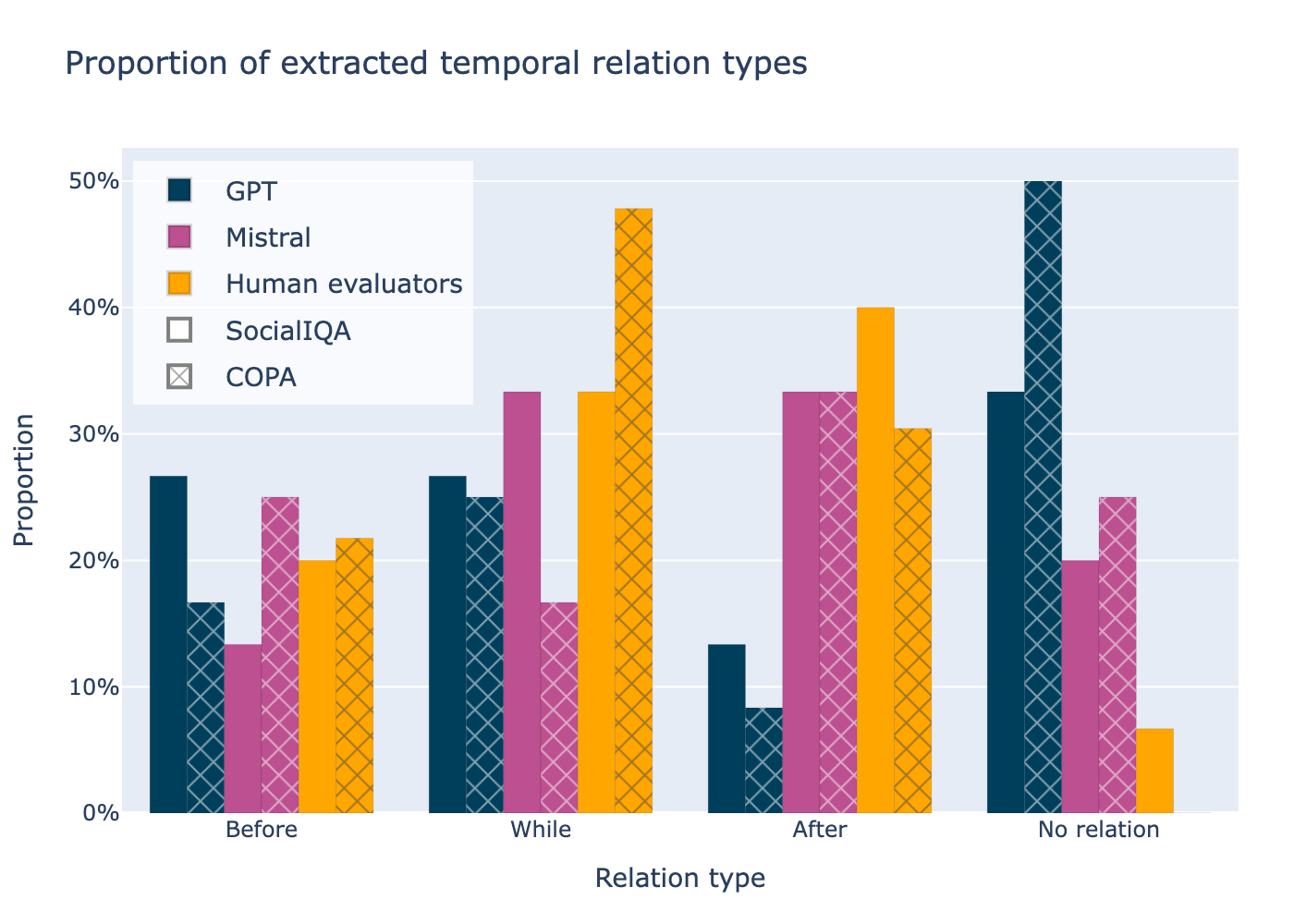}
    \caption{Relation type proportions for the temporal relationships extracted by models and humans.}
    \label{fig:time_rel_prop}
\end{figure}

\begin{figure}[H]
    \centering
    \includegraphics[width=0.75\linewidth]{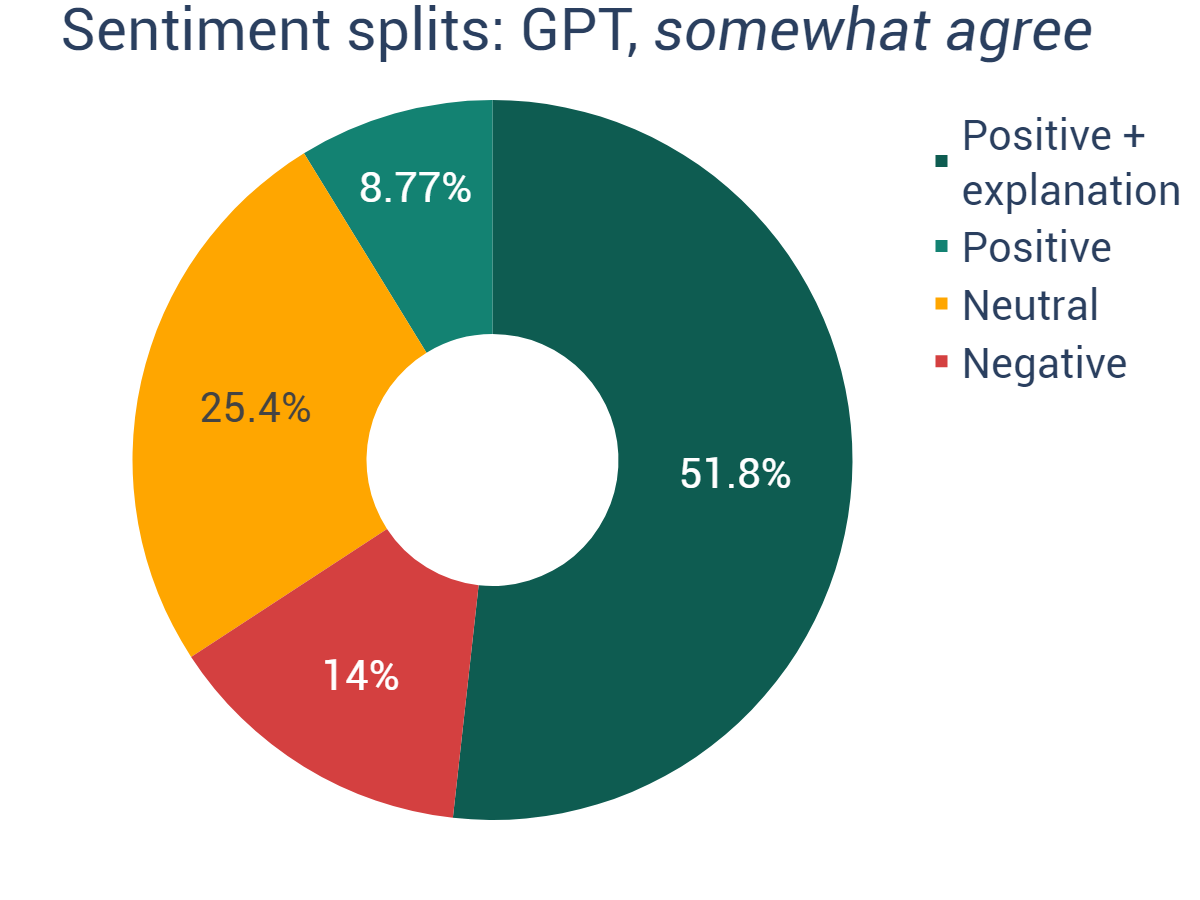}
    \includegraphics[width=0.75\linewidth]{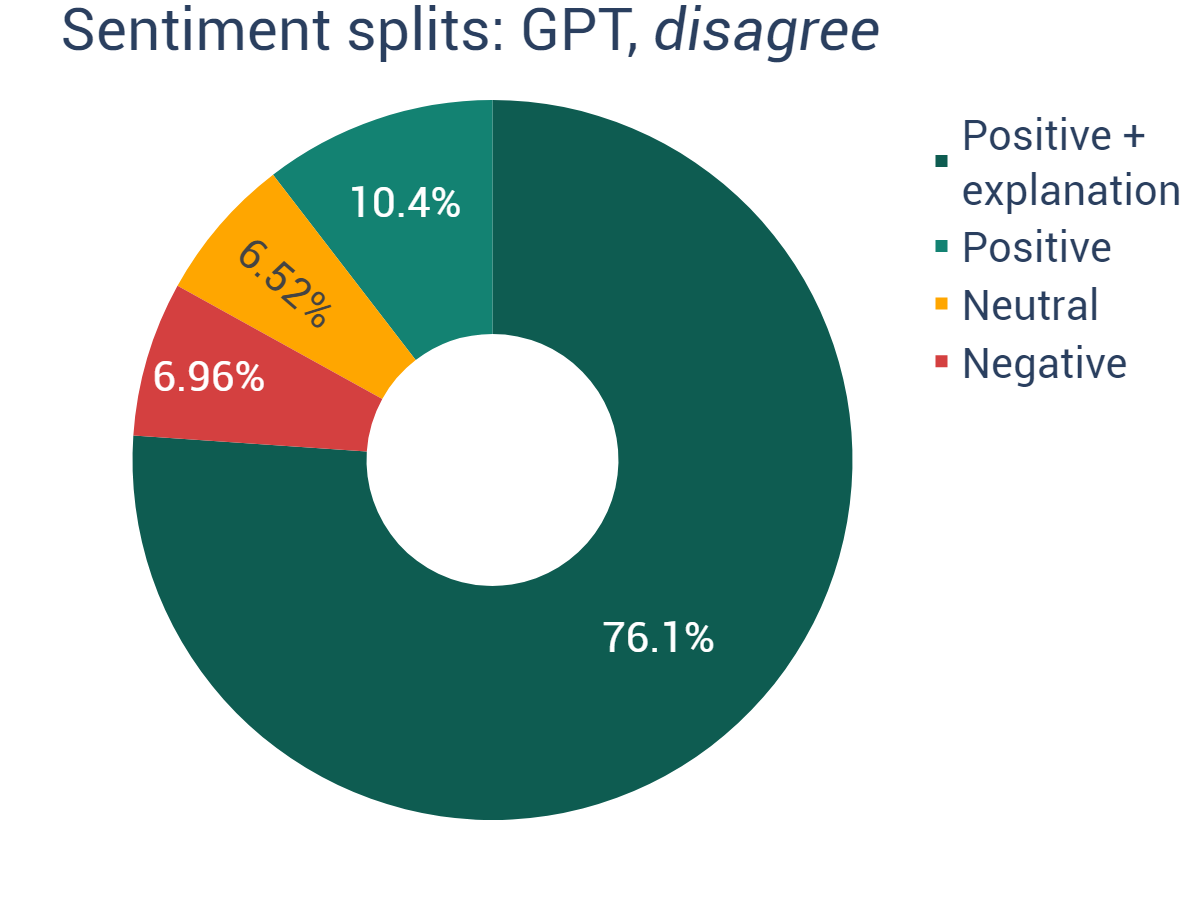}
    \includegraphics[width=0.75\linewidth]{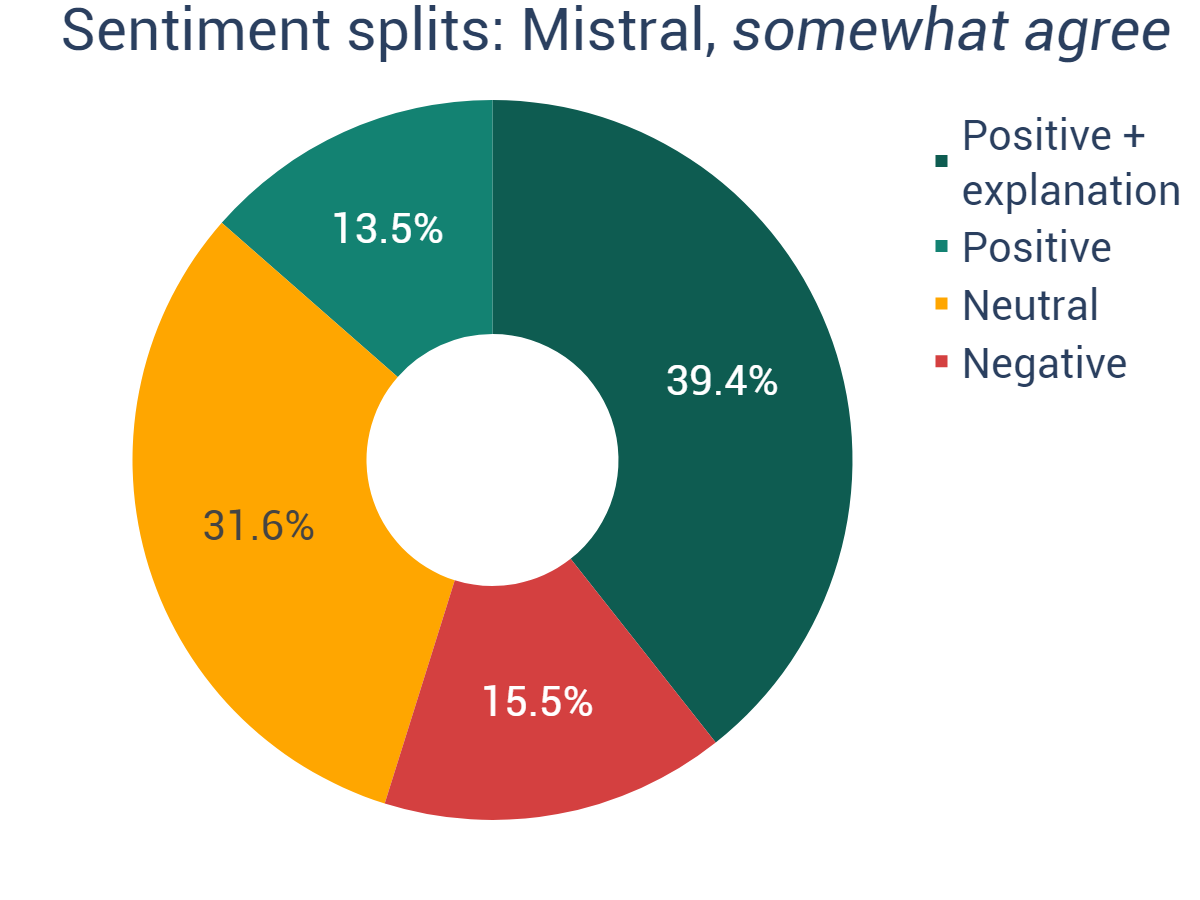}
    \includegraphics[width=0.75\linewidth]{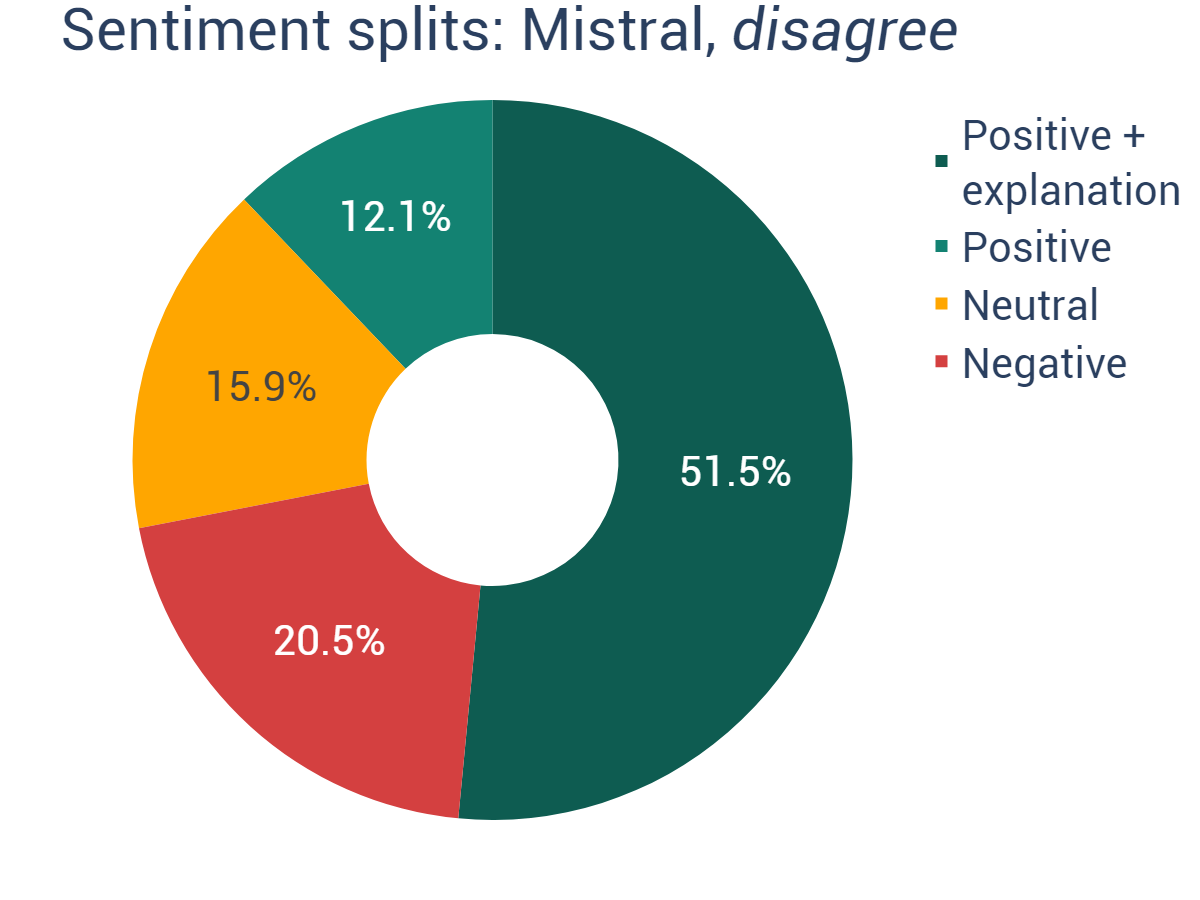}
    \caption{Human sentiment regarding discarded triplets in open answers for SocialIQA, grouped by model and multiple choice answer.}
    \label{fig:sentiment}
\end{figure}

Fig. \ref{fig:sentiment} reports human sentiment splits concerning discarded triplets as a function of the model generating the triplets as well as the answer choice expressing how strongly the evaluators agree with the decision to discard the triplet. Table \ref{table:inference_corr}, instead, details the results of the evaluation regarding the inference correction review section.

\begin{table*}[t]
\centering
\resizebox{\textwidth}{!}{%
\begin{tabular}{>{\raggedright\arraybackslash}l
                >{\raggedright\arraybackslash}l
                c c c c}
\hline
 & & \multicolumn{2}{c}{\textbf{SocialIQA}} & \multicolumn{2}{c}{\textbf{\rev{COPA}}} \\
\textbf{Sub-task} & \textbf{Metric} &
\textbf{Mistral} & \textbf{GPT} &
\textbf{\rev{Mistral}} & \textbf{\rev{GPT}} \\
\hline

& Model-Human Agreement
& 9 / 12 (75.0\%) & 5 / 10 (50.0\%)
& 12 / 12 (100\%) & 8 / 9 (88.9\%) \\

\rowcolor{black!10}\cellcolor{white}
& \textit{Fully agree} absolute majority rate
& 7 / 12 (58.3\%) & 4 / 10 (40.0\%)
& 11 / 12 (91.7\%) & 6 / 9 (66.7\%) \\

& \textit{Disagree} absolute majority rate
& 0 / 12 (0\%) & 5 / 10 (50.0\%)
& 0 / 12 (0\%) & 1 / 9 (11.1\%) \\

\rowcolor{black!10}\cellcolor{white}
& Average polarity
& 63.1\% & 45.6\%
& 87.6\% & 68.4\% \\

& Average \textit{Fully agree} proportion
& 49.9\% & 35.6\%
& 79.4\% & 55.6\% \\

\rowcolor{black!10}\cellcolor{white}\multirow{-6}{*}{Discard}
& Average \textit{Disagree} proportion
& 23.8\% & 44.4\%
& 4.2\% & 18.7\% \\

\hline

& Model-Human Agreement
& 12 / 12 (100\%) & 8 / 10 (80.0\%)
& 12 / 12 (100\%) & 8 / 9 (88.9\%) \\

\rowcolor{black!10}\cellcolor{white}\multirow{-2}{*}{Reason}
& Average polarity
& 86.3\% & 76.4\%
& 92.1\% & 78.0\% \\

\hline

& Model-Human Agreement
& 10 / 12 (83.3\%) & 10 / 10 (100\%)
& 12 / 12 (100\%) & 8 / 9 (88.9\%) \\

\rowcolor{black!10}\cellcolor{white}\multirow{-2}{*}{Correction}
& Average polarity
& 80.5\% & 77.6\%
& 89.8\% & 70.0\% \\

\hline
\end{tabular}%
}
\caption{Detailed results for the inference correction review section.}
\label{table:inference_corr}
\end{table*}

\subsubsection*{Timing Information}

Fig. \ref{fig:time_rel_prop} reports the relation type proportions for temporal relationships extracted by both models as well as humans.

\subsubsection*{Human Consensus}

\begin{figure}[ht]
    \centering
    \includegraphics[width=\linewidth]{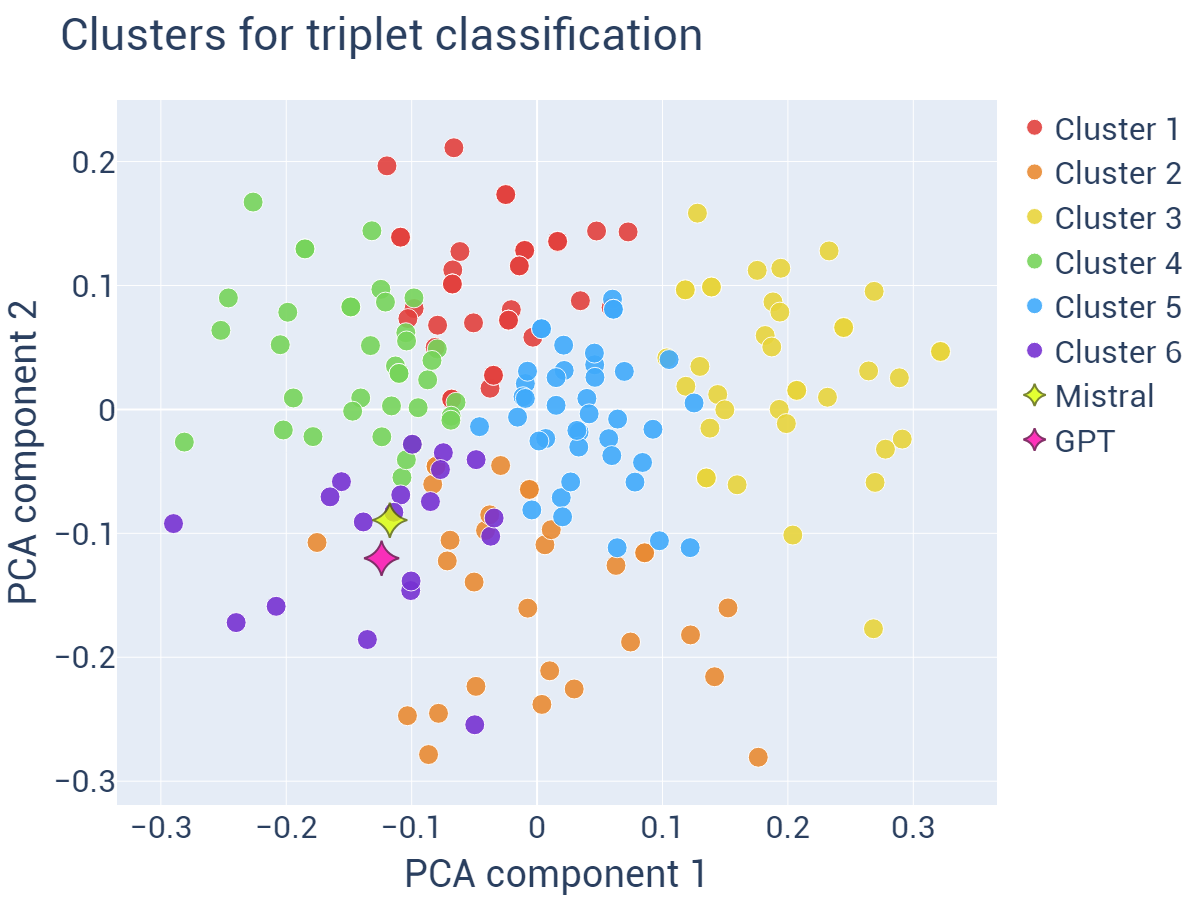}
    \caption{Clusters of respondents for the triplet classification task with markers highlighting the position of the two models.}
    \label{fig:cluster_class}
\end{figure}

\begin{figure}[ht]
    \centering
    \includegraphics[width=\linewidth]{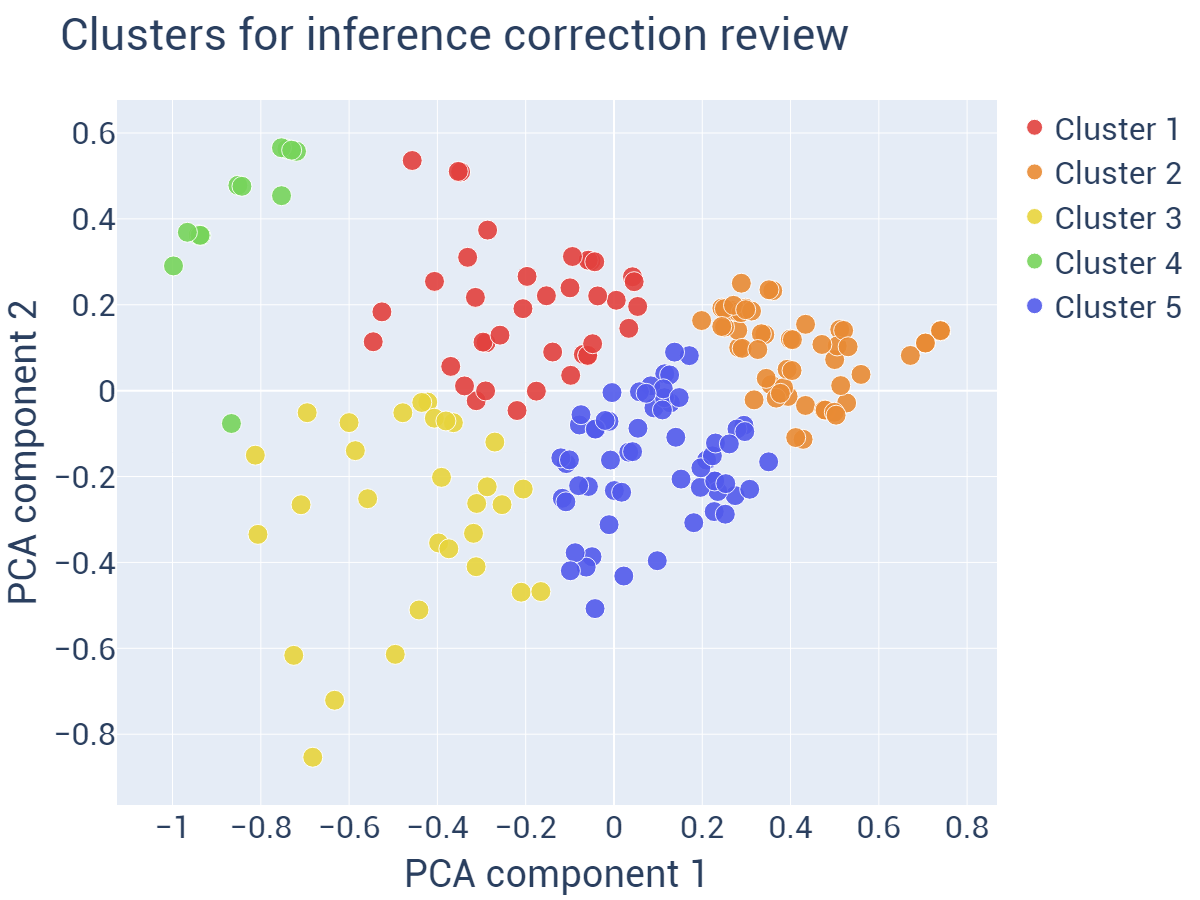}
    \caption{Clusters of respondents for the inference correction review task.}
    \label{fig:cluster_corr}
\end{figure}

\begin{figure}[ht]
    \centering
    \includegraphics[width=\linewidth]{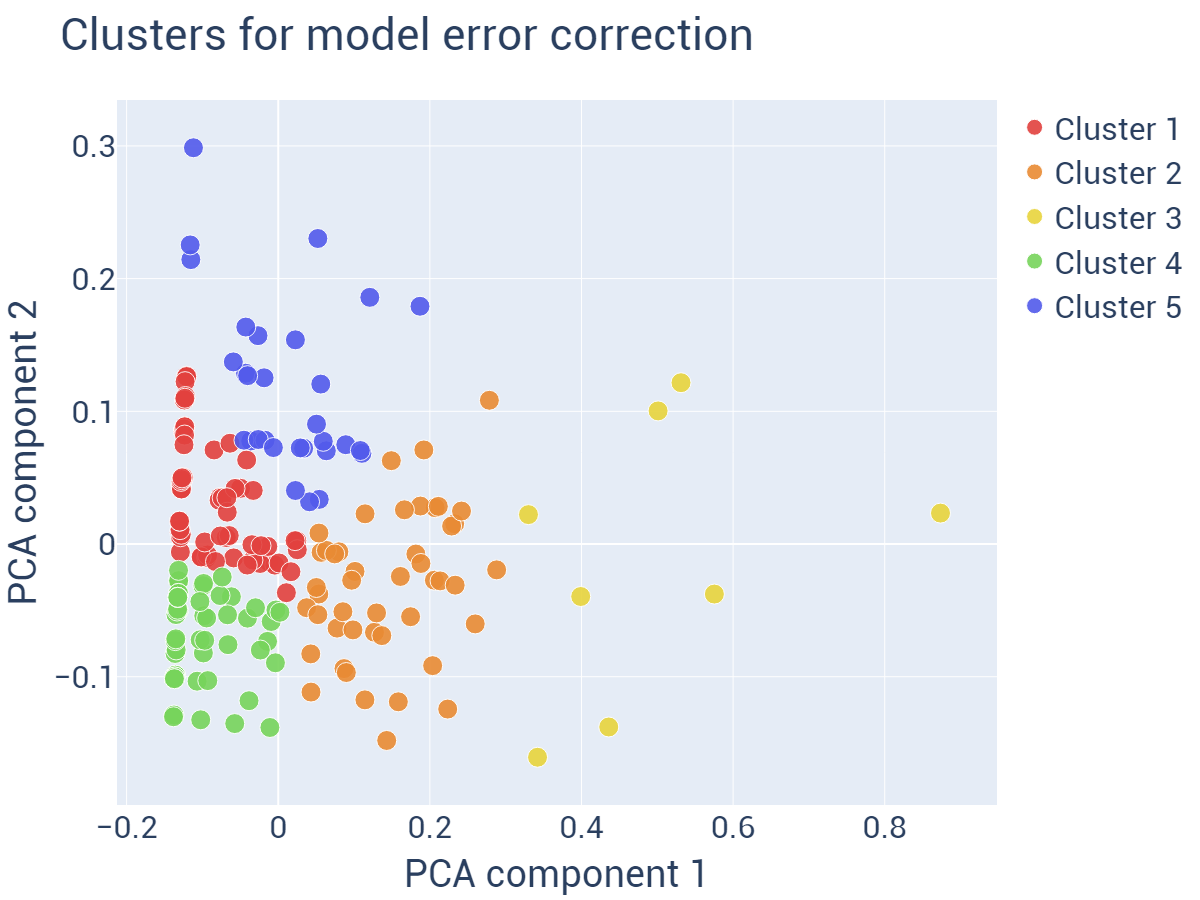}
    \caption{Clusters of respondents for the model error correction task.}
    \label{fig:cluster_err}
\end{figure}

Figures \ref{fig:cluster_class}, \ref{fig:cluster_corr}, and \ref{fig:cluster_err} show the cluster of human respondents for SocialIQA questions in triplet classification, inference correction review, and model error correction, respectively. Tables \ref{table:clust_class}, \ref{table:clust_corr}, and \ref{table:clust_err}, on the other hand, report the cluster-wise aggregated metrics referring to each of the clusters.

\begin{table*}[t]
    \small
    \centering 
    \begin{tabular}{>{\raggedright\arraybackslash}l >{\centering\arraybackslash}c >{\centering\arraybackslash}c >{\centering\arraybackslash}c >{\centering\arraybackslash}c >{\centering\arraybackslash}c >{\centering\arraybackslash}c >{\centering\arraybackslash}c}
    \hline
    \textbf{Metric}&\textbf{Cl. 1}&\textbf{Cl. 2}&\textbf{Cl. 3}&\textbf{Cl. 4}&\textbf{Cl. 5}&\textbf{Cl. 6}&\textbf{\textit{Average}}\\
    \hline
    \textit{Factual} proportion&39.2\%&38.7\%&\textbf{57.2\%}&32.2\%&46.2\%&35.2\%&\textit{42.2\%}\\
    \rowcolor{black!10}
    \textit{Deducible} proportion&38.3\%&24.8\%&21.8\%&\textbf{43.4\%}&31.6\%&39.1\%&\textit{33.0\%}\\
    \textit{Wrong} proportion&22.5\%&\textbf{36.5\%}&20.9\%&24.3\%&22.2\%&25.7\%&\textit{24.8\%}\\
    \rowcolor{black!10}
    Majority agreement rate&\textbf{92.5\%}&76.3\%&89.0\%&83.2\%&81.2\%&67.7\%&\textit{83.2\%}\\
    \hline
    \end{tabular}
    \caption{Cluster-wise aggregated metrics for triplet classification.}
    \label{table:clust_class}
\end{table*}

\begin{table*}[t]
    \small
    \centering 
    \begin{tabular}{>{\raggedright\arraybackslash}l >{\centering\arraybackslash}c >{\centering\arraybackslash}c >{\centering\arraybackslash}c >{\centering\arraybackslash}c >{\centering\arraybackslash}c >{\centering\arraybackslash}c}
    \hline
    \textbf{Metric}&\textbf{Cl. 1}&\textbf{Cl. 2}&\textbf{Cl. 3}&\textbf{Cl. 4}&\textbf{Cl. 5}&\textbf{\textit{Average}} \\
    \hline
    \textit{Fully agree} proportion&44.0\%&\textbf{84.9\%}&26.0\%&0.0\%&62.1\%&\textit{56.2\%}\\
    \rowcolor{black!10}
    \textit{Somewhat agree} proportion&6.2\%&3.6\%&\textbf{39.9\%}&2.6\%&24.5\%&\textit{15.7\%}\\
    \textit{Disagree} proportion&49.8\%&11.6\%&34.2\%&\textbf{97.4\%}&13.3\%&\textit{28.1\%}\\
    \rowcolor{black!10}
    Majority agreement rate&62.6\%&\textbf{83.8\%}&39.3\%&34.6\%&64.0\%&\textit{65.1\%}\\
    Open answer count&3.64&1.30&\textbf{5.83}&3.31&3.53&\textit{3.21}\\
    \rowcolor{black!10}
    Average open answer length&107.7&61.4&110.9&\textbf{124.0}&92.5&\textit{90.8}\\
    \hline
    \end{tabular}
    \caption{Cluster-wise aggregated metrics for inference correction review.}
    \label{table:clust_corr}
\end{table*}

\begin{table*}[t]
    \small
    \centering 
    \begin{tabular}{>{\raggedright\arraybackslash}l >{\centering\arraybackslash}c >{\centering\arraybackslash}c >{\centering\arraybackslash}c >{\centering\arraybackslash}c >{\centering\arraybackslash}c >{\centering\arraybackslash}c}
    \hline
    \textbf{Metric}&\textbf{Cl. 1}&\textbf{Cl. 2}&\textbf{Cl. 3}&\textbf{Cl. 4}&\textbf{Cl. 5}&\textbf{\textit{Average}}\\
    \hline
    Triplet discard rate&34.0\%&31.1\%&32.7\%&25.4\%&\textbf{39.2\%}&\textit{31.7\%}\\
    \rowcolor{black!10}
    Majority agreement rate&87.3\%&91.3\%&87.1\%&\textbf{93.2\%}&79.4\%&\textit{88.7\%}\\
    Average addition length&6.03&36.61&\textbf{83.30}&4.59&18.09&\textit{17.29}\\
    \hline
    \end{tabular}
    \caption{Cluster-wise aggregated metrics for model error correction.}
    \label{table:clust_err}
\end{table*}

\end{document}